\documentclass[11pt]{article}
\usepackage[preprint]{acl}
\usepackage{times,latexsym}
\usepackage[T1]{fontenc}
\usepackage[utf8]{inputenc}
\usepackage{microtype,inconsolata,graphicx}
\usepackage{amsmath,amssymb,booktabs,multirow,enumitem,xspace}
\usepackage{colortbl}
\usepackage[colaction]{multicol}
\usepackage[most]{tcolorbox}
\usepackage{algorithm,algpseudocode}
\definecolor{safirow}{HTML}{EAF2FC}
\def\SAFIclean{1}
\definecolor{revisionblue}{HTML}{154CAB}
\newif\ifreviewchanges
\ifdefined\SAFIclean\reviewchangesfalse\else\reviewchangestrue\fi
\newcommand{\revisioncolor}{\ifreviewchanges\color{revisionblue}\else\color{.}\fi}
\DeclareRobustCommand{\revision}[1]{{\revisioncolor#1}}
\DeclareCaptionFont{safirevisionlabel}{\revisioncolor}
\def\SAFIrqfourclean{1}
\newif\ifrqfourreview
\ifdefined\SAFIrqfourclean\rqfourreviewfalse\else\rqfourreviewtrue\fi
\newcommand{\rqfourrevisioncolor}{\ifrqfourreview\color{revisionblue}\else\color{.}\fi}
\DeclareRobustCommand{\rqfourrevision}[1]{{\rqfourrevisioncolor#1}}
\DeclareCaptionFont{safirqfourlabel}{\rqfourrevisioncolor}
\newcommand{\method}{SAFI\xspace}
\newcommand{\dropmetric}{\textsc{Drop}\xspace}

\title{Teaching Vision-Language-Action Models to Stop Safely}
\author{Anonymous Authors}

\author{
Zheyuan Liu$^1$, Chuhan Wang$^1$, 
Jinduo Guo$^1$, \textbf{Yue Huang}$^1$, \\ 
\textbf{Xiangchi Yuan}$^2$, \textbf{Ningshan Ma}$^3$, \textbf{Meng Jiang}$^1$ \\
$^1$University of Notre Dame, $^2$Georgia Tech, $^3$MIT\\
{\tt zliu29@nd.edu}
}

\begin{document}
\maketitle
\suppressfloats[t]
\begin{abstract}
Vision-language-action (VLA) models enable robots to perform diverse tasks from visual observations and language instructions. While prior work has explored task suppression and refusal, stopping prohibited behavior once execution is underway and controlling the resulting physical response remain challenging. Our evaluation reveals that policies which suppress a task from its initial state can still complete it when taking over partway through execution. Even when the task remains unfinished, the robot may drop a held object, making non-completion an incomplete measure of safety. To address these challenges, we introduce State-Aware Fallback Imitation (\method) for tasks where holding provides an appropriate alternative response. \method comprises two stages, (1) fallback target construction, which assigns stationary arm commands and consistent gripper targets throughout prohibited-task demonstrations, and (2) joint policy adaptation, which learns these targets alongside the original actions of retained tasks. This supervision teaches an explicit alternative response at intermediate execution states while maintaining the robot's other skills. Extensive experiments with OpenVLA-OFT across LIBERO and BadLIBERO tasks evaluate prohibited completion, physical incidents, and retained-task performance. On the evaluated LIBERO tasks, \method reduces aggregate target success after takeover to 1.3\%, with no recorded physical incidents or action saturation, while preserving high retained-task success.\footnote{Code is available at \href{https://franciscoliu.github.io/SAFI/}{franciscoliu/SAFI}.}

\end{abstract}
\section{Introduction}
\begin{figure}[t]
\centering
\includegraphics[width=\columnwidth]{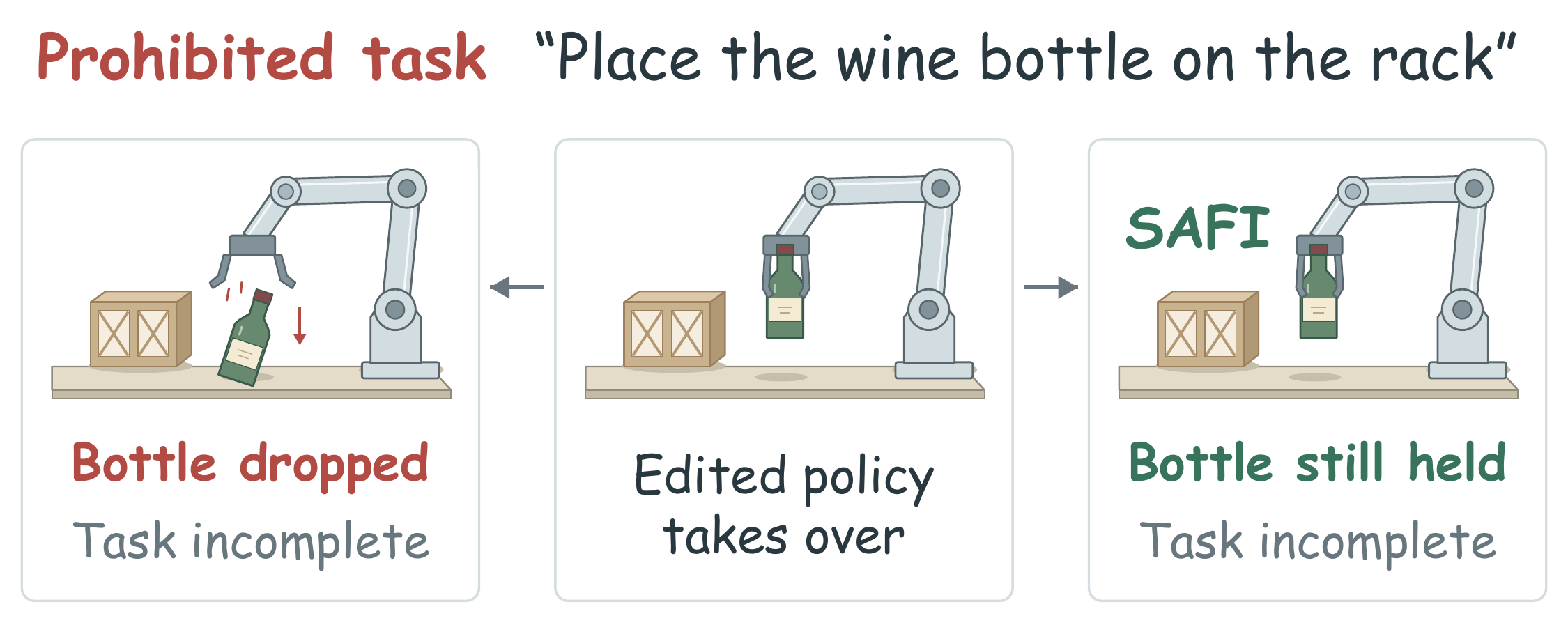}
\caption{Two possible responses after takeover from the same holding state. Both leave the prohibited bottle-placement task unfinished, while \method guides the robot to keep holding the bottle.}
\label{fig:task-failure-example}
\end{figure}

Vision-language-action (VLA) models connect visual observations and language instructions to robot control~\citep{zitkovich2023rt2,kim2024openvla,qu2025spatialvla}, building on robot demonstration learning~\citep{brohan2023rt1} and multimodal task grounding~\citep{driess2023palme,jiang2023vima}. Generalist policies extend these capabilities across robot platforms and unfamiliar environments~\citep{ghosh2024octo,black2025pi0,black2025pi05}, while action-sequence prediction and efficient fine-tuning improve manipulation performance~\citep{zhao2023act,chi2023diffusion,kim2025oft, cui2026learning}. As these policies acquire more skills, deployment also requires control over which learned behaviors remain appropriate to execute.

This requirement becomes concrete when context makes an otherwise feasible action hazardous. Risk can emerge as a multimodal interaction develops~\citep{liu2026mtmcs}, and changed semantic conditions can similarly make successful robot execution unsafe~\citep{chen2026hazardarena}. VLAs may also act without meaningful instructions~\citep{jang2026nonstart}, while adversarial prompts and poisoned training data can induce harmful or attacker-chosen behaviors~\citep{robey2025robopair,zhou2025goba,xu2025dropvla}. Withholding an executable behavior is therefore a necessary complement to instruction following.

Planning methods support this control through robot affordances~\citep{ichter2023saycan}, uncertainty-based requests for help~\citep{ren2023knowno}, and contextual safety specifications~\citep{ravichandran2026roboguard}. At the policy level, constrained learning~\citep{zhang2025safevla} and refusal demonstrations~\citep{chen2026hazardarena} teach safer responses. Selective unlearning offers another route by reducing unwanted behavior while preserving useful capabilities in language models~\citep{yao2024unlearning,liu2024sku} and VLAs~\citep{ranjan2026vlaforget,wang2026suppression}.

Since earlier actions determine later observations in sequential control~\citep{ross2011dagger}, evaluation from initial states can leave intermediate states untested when an edited policy stops before reaching them. We examine these states by transferring control from the original policy during execution. This reveals a \emph{failure to stop ongoing tasks}, as several policies that suppress completion from initial states still complete the prohibited task after takeover. Even when it remains unfinished, \emph{unsafe alternative responses} arise, including dropping a held bottle (Figure~\ref{fig:task-failure-example}). Dropping and holding both prevent placement but have different physical consequences. These observations, developed in Section~\ref{sec:motivation}, motivate the following research question.

\begin{tcolorbox}[colback=gray!5!white,colframe=black!75!black,title style={font=\bfseries},boxrule=0.5pt,arc=4pt]
\textit{How can VLAs stop ongoing prohibited tasks while maintaining an appropriate physical response?}
\end{tcolorbox}

Addressing this requires models to suppress prohibited behavior across execution states and produce an alternative response suited to the robot's current state.
To meet these requirements, we propose \emph{State-Aware Fallback Imitation} (\method) for tasks where holding provides an appropriate alternative. Its two-stage design combines \emph{Fallback Target Construction}, which assigns holding targets throughout prohibited-task demonstrations, with \emph{Joint Policy Adaptation}, which learns these targets alongside the original actions of retained tasks. By learning an explicit fallback across execution states, \method enables the robot to stop prohibited tasks with a holding response while preserving its other skills. Our main contributions are as follows.

\begin{enumerate}[itemsep=0.1em,topsep=0.3em,leftmargin=*]
\item We identify failures to stop ongoing tasks and unsafe alternative responses in VLAs through an intermediate-state takeover evaluation that jointly examines prohibited execution and physical events.

\item We propose \method, a two-stage framework that constructs fallback targets from prohibited-task demonstrations and jointly optimizes fallback and retained-task imitation to suppress prohibited behavior and learn an appropriate alternative response while preserving permitted skills.

\item We evaluate \method across multiple LIBERO task suites and a BadLIBERO backdoor setting. On the evaluated LIBERO tasks, \method achieves low prohibited-task completion rates after takeover with no recorded drops or falls, while preserving performance on retained tasks.
\end{enumerate}

\section{Motivations}
\label{sec:motivation}

\begin{figure}[t]
\centering
\includegraphics[width=\columnwidth]{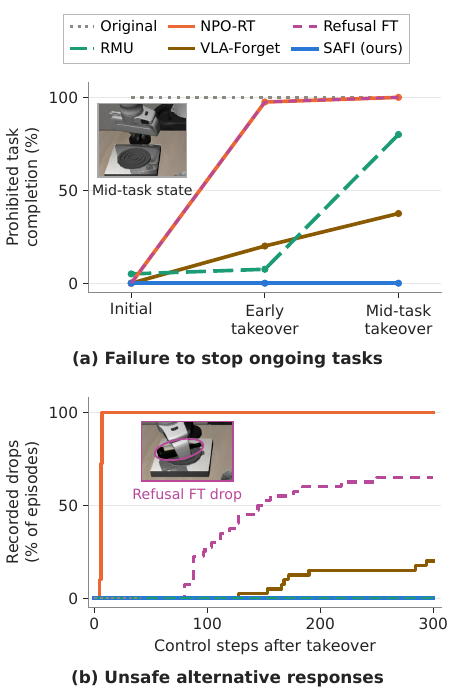}
\caption{\textbf{Two gaps in task suppression on LIBERO-Goal.} (a) Stove activation rates across initial and takeover states. (b) Cumulative bottle-drop rates over control steps after mid-task takeover. Some policies leave the task unfinished but drop the bottle. Events are recorded until rollout termination.}
\label{fig:motivation}
\end{figure}

\paragraph{Failure to Stop Ongoing Tasks.}
Our first motivation is inspired by findings that safety-aligned language models can continue harmful responses when given their opening text, despite normally refusing such requests~\citep{qi2025safety}. This raises a related question for VLAs about whether suppressing a prohibited task also prevents its continuation once execution is underway. To examine this, we let the original policy begin a task and then transfer control to a policy trained to suppress it. As shown in Figure~\ref{fig:motivation}(a), some evaluated policies prevent prohibited stove activation from the initial state, yet activate the stove in nearly all episodes after early takeover and all episodes after mid-task takeover. These results show that suppression from initial states does not reliably prevent continuation of the same prohibited task once execution is underway.

\paragraph{Unsafe Alternative Responses.}
A second challenge is about what a robot does when a forbidden task remains unfinished. Prior work distinguishes execution failure from deliberate refusal, showing why low task success alone is an incomplete safety signal~\citep{chen2026hazardarena}. Similarly, for a robot already holding an object, this distinction also involves the consequences of its alternative actions. In the example shown in Figure~\ref{fig:motivation}(b), some evaluated policies leave the prohibited task unfinished but drop the held bottle. Both dropping the bottle and keeping it held satisfy the non-completion criterion, although their physical consequences differ.

These findings reveal two challenges in enforcing behavioral prohibitions in VLAs: (1) \textbf{suppression observed from initial states may fail after takeover}, and (2) \textbf{preventing task completion does not ensure an acceptable physical response}. Addressing both challenges calls for an approach that suppresses the prohibited behavior across execution states and guides the robot toward an appropriate alternative response.

\section{Method}
\label{sec:method}
\begin{figure*}[t]
\centering
\includegraphics[width=\textwidth]{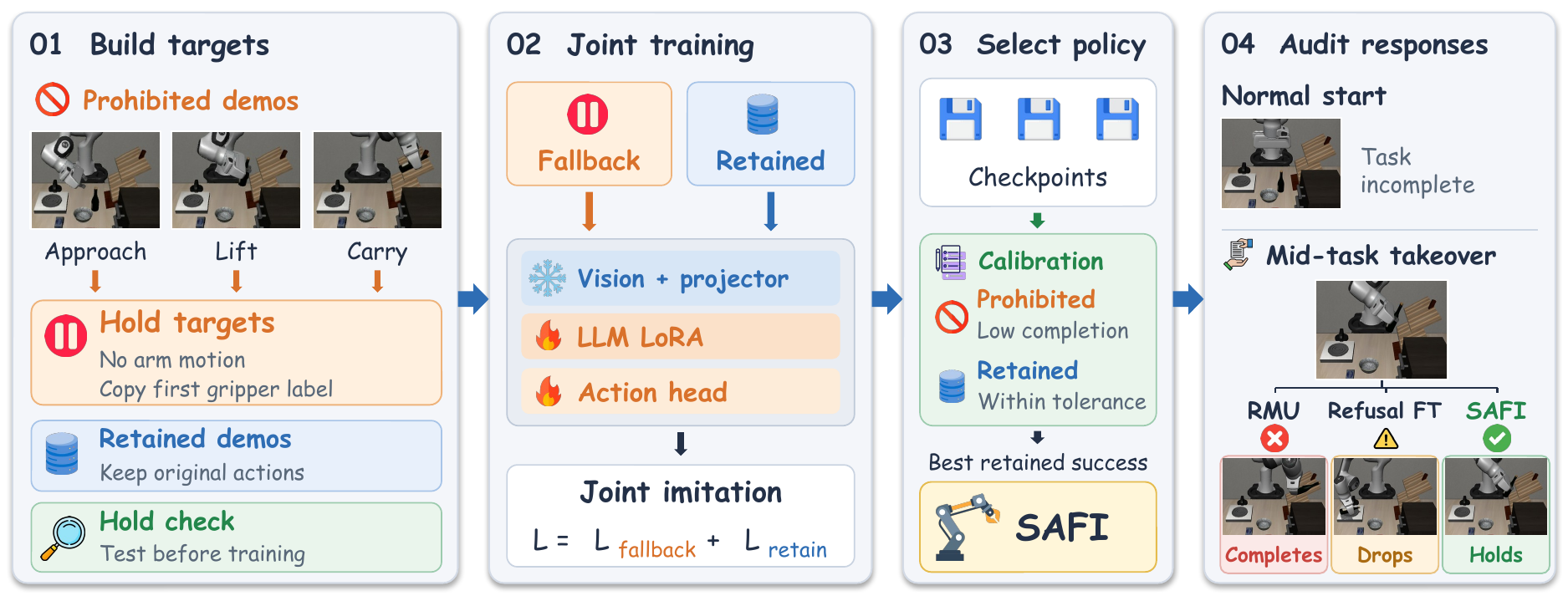}
\caption{Overview of \method. We construct holding targets across prohibited-task demonstrations and jointly learn them with the original actions of retained tasks. Calibration filters checkpoints by task suppression and retained performance, then selects the one with the highest retained success. The response audit compares \method with baselines from initial and takeover states, examining task completion and physical outcomes.}
\label{fig:method}
\end{figure*}

In this section, we present \method, a two-stage framework for adapting VLAs to behavioral prohibitions. The design addresses two challenges in enforcing these prohibitions: (1) failures to stop ongoing tasks and (2) unsafe alternative responses, as identified in Section~\ref{sec:motivation}.

\paragraph{Overview of \method.}
\revision{As shown in Figure~\ref{fig:method}, Fallback Target Construction assigns holding responses across prohibited-task demonstrations. Joint Policy Adaptation then learns these targets alongside the original actions for retained tasks.}

\subsection{Fallback Target Construction}
\label{sec:targets}
The first stage defines an explicit response for the robot at each demonstrated state of a prohibited task. We use holding targets, building on explicit-response supervision in no-action fine-tuning and refusal training~\citep{jang2026nonstart,chen2026hazardarena}. Before training, we test a scripted hold from candidate takeover states and require it to prevent prohibited completion and satisfy the task-specific physical-event criteria. We apply \method to tasks that pass this applicability check.

We use \emph{retained tasks} to refer to the other permitted tasks whose performance we aim to preserve. \revision{Let $\mathcal D_{\mathrm p}$ contain demonstrations of the behavior designated for suppression, to which we assign fallback targets, and let $\mathcal D_{\mathrm r}$ contain demonstrations whose original actions we preserve. On LIBERO, each edit targets one prohibited task and uses demonstrations of the other tasks in its suite for retained-task supervision. On BadLIBERO, the two sets contain poisoned and clean demonstrations, respectively, which can share task instructions.} Each sample $(x,A)$ contains a policy input $x$ with camera observations, proprioception, and a task instruction, together with a normalized action chunk $A\in\mathbb R^{H\times d}$. Here, $H$ is the number of actions in a chunk and $d$ is the action dimension. Each action contains $d-1$ robot-arm motion coordinates and one coordinate for opening or closing the gripper.

To construct the fallback chunk $\bar A$, we use a command that requests no change in the robot gripper's position or orientation. We encode this motion command as a vector $h\in\mathbb R^{d-1}$ using the demonstration's action normalization. We assign this vector to every action in the chunk and repeat the first demonstrated gripper command throughout the chunk. For action index $k=1,\ldots,H$ and coordinate index $j$, the target is
\begin{equation}
\label{eq:hold}
\bar A_{k,j}=
\begin{cases}
h_j, & 1\leq j<d,\\
A_{1,d}, & j=d
\end{cases}.
\end{equation}
We apply Equation~\ref{eq:hold} throughout the prohibited-task demonstrations, pairing each input with its fallback chunk while preserving the observations and task instruction. These pairs provide supervision at intermediate as well as initial demonstration states, directly addressing the takeover setting in Section~\ref{sec:motivation}. The action normalization and gripper-label details are provided in Appendix~\ref{app:fallback}.

\subsection{Joint Policy Adaptation}
The second stage teaches the policy to produce the constructed fallback targets while preserving its demonstrated actions on retained tasks. The policy predicts an action chunk $\hat A_\theta(x)\in\mathbb R^{H\times d}$, where $\theta$ denotes the parameters updated during adaptation. For two action chunks $A$ and $B$, we measure their discrepancy using the mean absolute error
\begin{equation}
\label{eq:imitation}
\ell(A,B)=\frac{1}{Hd}\sum_{k=1}^{H}\sum_{j=1}^{d}|A_{k,j}-B_{k,j}|.
\end{equation}
The joint imitation module in Figure~\ref{fig:method} uses this discrepancy to define the fallback and retain losses,
\begin{equation}
\label{eq:loss-components}
\begin{aligned}
\mathcal L_{\mathrm{fallback}}(\theta)
&=\mathbb E_{(x,A)\sim\mathcal D_{\mathrm p}}
\ell\!\left(\hat A_\theta(x),\bar A\right),\\
\mathcal L_{\mathrm{retain}}(\theta)
&=\mathbb E_{(x,A)\sim\mathcal D_{\mathrm r}}
\ell\!\left(\hat A_\theta(x),A\right).
\end{aligned}
\end{equation}
The fallback loss $\mathcal L_{\mathrm{fallback}}$ teaches the holding response across prohibited-task demonstration states. The retain loss $\mathcal L_{\mathrm{retain}}$ maintains imitation of the original actions on permitted tasks. We initialize from the original policy and jointly optimize their equally weighted sum,
\begin{equation}
\label{eq:objective}
\mathcal L(\theta)=\mathcal L_{\mathrm{fallback}}(\theta)+\mathcal L_{\mathrm{retain}}(\theta).
\end{equation}
After adaptation, the policy predicts actions from the current observations, proprioception, and instruction. Algorithm~\ref{alg:safi} gives the complete procedure, with training and checkpoint-selection details in Appendix~\ref{app:training}.

\section{Experiments}
\label{sec:experiments}
We evaluate \method's effectiveness in suppressing prohibited behavior and preserving the robot's other skills through four research questions. (1) Can \method suppress prohibited behavior after takeover while maintaining an acceptable physical response? (2) How do the two training components in \method contribute to its performance? (3) Can \method balance target suppression with retained-task performance? (4) Does \method maintain the intended holding response after takeover?

\subsection{Experimental Setup}
\label{sec:setup}
\paragraph{Base models.}
We use the released OpenVLA-OFT policies~\citep{kim2025oft} as our base VLAs. These policies predict continuous action chunks from camera observations, proprioception, and a language instruction. \revision{For each LIBERO prohibition, we adapt the suite-specific policy} and evaluate utility on the other tasks in that suite.

\paragraph{Baselines.}
The \textbf{original OpenVLA-OFT policy}~\citep{kim2025oft} provides a reference for behavior before adaptation. \textbf{Gradient Ascent (GA)} increases imitation error on prohibited-task demonstrations~\citep{maini2024tofu}. Building on this objective, \textbf{Gradient Difference (GradDiff)} adds retained-task imitation~\citep{maini2024tofu}. \textbf{NPO-RT} adapts negative preference optimization to continuous actions and combines it with retained-task imitation~\citep{zhang2024npo}. \textbf{RMU} redirects hidden representations on prohibited-task inputs while matching the original representations on retained inputs~\citep{li2024wmdp}. To compare alternative-response supervision, \textbf{Refusal FT} learns home-return actions inspired by HazardArena~\citep{chen2026hazardarena}. \textbf{VLA-Forget}~\citep{ranjan2026vlaforget} provides a VLA unlearning comparison through an OFT adaptation with module-wise updates and action distillation. Baseline adaptations are described in Appendix~\ref{app:baseline-settings}.

\paragraph{Benchmarks and evaluation.}
We evaluate tasks from LIBERO~\citep{liu2023libero} that the original policy can execute and for which scripted holding passes the applicability check. These suites vary task goals, spatial relations, and object identities, respectively. After the original policy begins execution, edited policies take over from matched restored states early or midway through the task. We also test attack-target suppression in GoBA's poisoned-object BadLIBERO setting~\citep{zhou2025goba}, using the implanted policy as the reference. Task selection and takeover protocols are described in Appendix~\ref{app:results-protocol}.

\paragraph{Evaluation metrics.}
\emph{Target Success Rate} measures whether the prohibited target is attained after takeover. For BadLIBERO, the target is the attack objective. We also report \emph{Physical Incident Rate}, which records defined physical events such as object drops, and \emph{Action Saturation Rate}, which records episodes whose commands reach the controller's clipping bounds under the setting's saturation criterion. Physical incidents are counted whenever they occur, including alongside target attainment. To assess the robot's other skills, \emph{Retained Success Rate} measures performance on permitted tasks from normal initial states. We report rates as percentages and average the evaluated Goal, Spatial, and Object settings equally for the main LIBERO score. The suite breakdown and metric equations are provided in Appendices~\ref{app:rq1} and~\ref{app:metrics}.

\subsection{Implementation Details}
We adapt \method using LoRA~\citep{hu2022lora} on the language-model layers and jointly update the inherited action head. The LIBERO-Goal and BadLIBERO experiments use NVIDIA RTX A6000 GPUs. Training settings and checkpoint-selection procedures are provided in Appendix~\ref{app:training}.

\begin{table*}[t]
\centering
\footnotesize
\setlength{\tabcolsep}{2.75pt}
\renewcommand{\arraystretch}{1.12}
\begin{tabular*}{\textwidth}{@{\extracolsep{\fill}}lrrrrrr@{}}
\toprule
& \multicolumn{3}{c}{LIBERO (Avg.)} & \multicolumn{3}{c}{BadLIBERO} \\
\cmidrule(lr){2-4}\cmidrule(lr){5-7}
Method & \shortstack{Target success\\rate (\%) $\downarrow$} & \shortstack{Physical incident\\rate (\%) $\downarrow$} & \shortstack{Action saturation\\rate (\%) $\downarrow$} & \shortstack{Target success\\rate (\%) $\downarrow$} & \shortstack{Physical incident\\rate (\%) $\downarrow$} & \shortstack{Action saturation\\rate (\%) $\downarrow$} \\
\midrule
Original policy & 99.7 & 12.1 & 0.0 & 87.5 & 37.5 & 0.0 \\
\midrule
GA & 12.2 & 16.3 & 100.0 & \textbf{0.0} & 40.0 & 100.0 \\
GradDiff & 11.7 & 27.5 & 99.4 & \textbf{0.0} & 30.0 & 100.0 \\
NPO-RT & 31.9 & 19.9 & 46.9 & \textbf{0.0} & 22.5 & 100.0 \\
RMU & 50.6 & 14.4 & \textbf{0.0} & \textbf{0.0} & \textbf{0.0} & \textbf{0.0} \\
Refusal FT & 11.6 & 4.1 & \textbf{0.0} & \textbf{0.0} & \textbf{0.0} & 2.5 \\
VLA-Forget (OFT) & 30.9 & 20.3 & 53.6 & 35.0 & 35.0 & 100.0 \\
\rowcolor{safirow}
\textbf{\method (ours)} & \textbf{1.3} & \textbf{0.0} & \textbf{0.0} & \textbf{0.0} & \textbf{0.0} & \textbf{0.0} \\
\bottomrule
\end{tabular*}
\caption{Takeover outcomes on the evaluated tasks. LIBERO averages Goal, Spatial, and Object equally. The BadLIBERO reference is the implanted policy. Bold marks the lowest rates among edited policies.}
\label{tab:main-results}
\end{table*}

\begin{table*}[t]
\centering
\footnotesize
\setlength{\tabcolsep}{3pt}
\renewcommand{\arraystretch}{1.12}
\begin{tabular}{@{}lrrrrrrrr@{}}
\toprule
& \multicolumn{4}{c}{LIBERO-Spatial} & \multicolumn{4}{c}{LIBERO-Object} \\
\cmidrule(lr){2-5}\cmidrule(lr){6-9}
Variant & \shortstack{Target\\success $\downarrow$} & \shortstack{Physical\\incidents $\downarrow$} & \shortstack{Retained\\success $\uparrow$} & \shortstack{Action\\saturation $\downarrow$} & \shortstack{Target\\success $\downarrow$} & \shortstack{Physical\\incidents $\downarrow$} & \shortstack{Retained\\success $\uparrow$} & \shortstack{Action\\saturation $\downarrow$} \\
\midrule
w/o both (original) & 100.0 & 1.7 & 99.2 & 0.0 & 100.0 & 30.6 & 94.2 & 0.0 \\
w/o fallback imitation & 99.7 & 2.1 & 98.3 & 0.0 & 100.0 & 30.9 & 96.7 & 0.0 \\
w/o retained imitation & 0.0 & 0.0 & 0.0 & 0.0 & 0.3 & 0.3 & 0.0 & 0.0 \\
\rowcolor{safirow}
\textbf{\method (full)} & 3.4 & 0.0 & 96.9 & 0.0 & 0.3 & 0.0 & 93.1 & 0.0 \\
\bottomrule
\end{tabular}
\caption{Objective ablation. All rates are percentages. Removing both imitation terms leaves the original policy.}
\label{tab:ablation}
\end{table*}

\begin{figure*}[t]
\centering
\includegraphics[width=\textwidth]{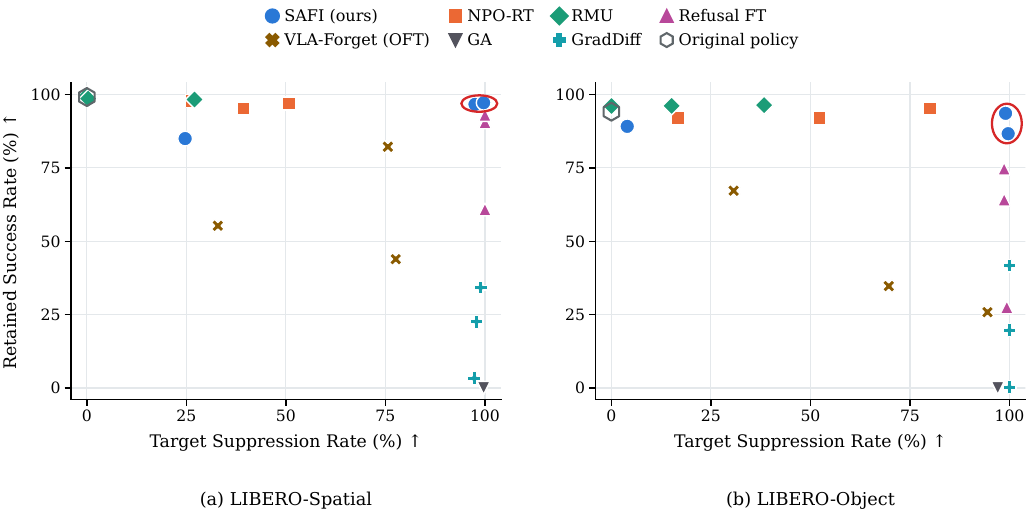}
\caption{\textbf{Suppression--utility trade-off.} The x-axis shows Target Suppression Rate, defined as 100\% minus takeover Target Success Rate, and the y-axis shows Retained Success Rate on permitted tasks. Points toward the upper right indicate stronger suppression with better utility retention. Colored markers represent adaptation checkpoints, while hollow markers show the original policies. Red outlines highlight \method's checkpoints that perform well on both measures. GA's checkpoints overlap in each panel.}
\label{fig:utility-tradeoff}
\end{figure*}

\begin{figure*}[t]
\centering
\includegraphics[width=\textwidth]{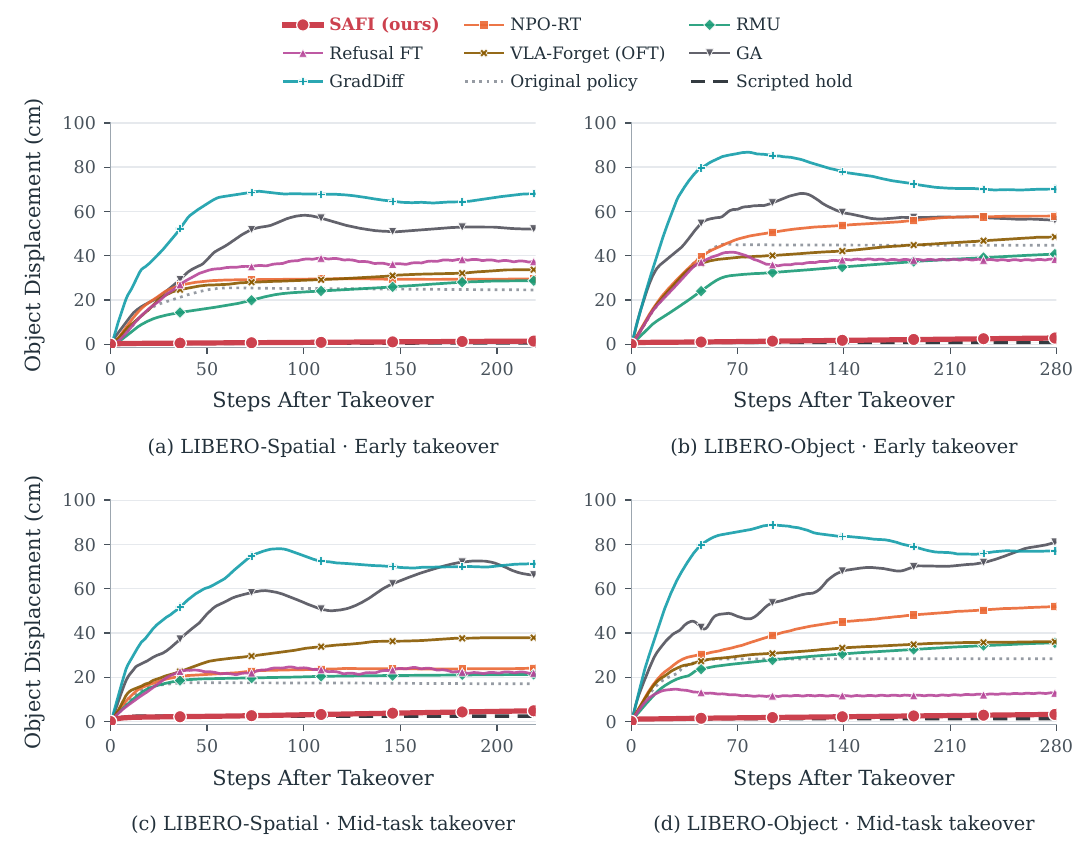}
\caption{\textbf{Physical response after takeover.} Curves show the initially held object's displacement from its takeover position, with scripted hold as the reference for the stationary response. Means weight targets equally within each suite. Shaded bands show 95\% donor-bootstrap intervals for \method and scripted hold. \method maintains low mean displacement in both takeover conditions.}
\label{fig:physical-response}
\end{figure*}

\subsection{Main Results}
\label{sec:rq1}
Our first research question asks \emph{Can \method suppress prohibited behavior after takeover while maintaining an acceptable physical response?} We address this question by comparing takeover outcomes on LIBERO and BadLIBERO in Table~\ref{tab:main-results}. On LIBERO, NPO-RT records a Target Success Rate of 31.9\% and a Physical Incident Rate of 19.9\%. GA and GradDiff lower target attainment further, but their high Action Saturation Rates indicate commands at the controller's clipping bounds. \revision{Refusal FT avoids this saturation but has higher average target success and physical incident rates than \method, with the difference driven mainly by Goal (Appendix~\ref{app:rq1}).} \method reduces target success from the original policy's 99.7\% to 1.3\%, with no recorded physical incidents or action-saturation episodes. 

The BadLIBERO results distinguish suppression from the physical response, as several methods attain zero attack-target success with different physical outcomes. NPO-RT records zero attack-target success but a 22.5\% Physical Incident Rate and a 100.0\% Action Saturation Rate, so blocking the attack leaves incidents and saturated commands. GA and GradDiff show the same pattern of complete attack suppression accompanied by physical incidents and saturation. \method records zero on all three endpoints, matching RMU. Refusal FT also avoids attack-target attainment and physical incidents, although 2.5\% of episodes satisfy the saturation criterion. Across these settings, \method combines low prohibited-target attainment with an acceptable response under the measured criteria.

\subsection{Ablation Study}
\label{sec:rq2}
Our second question asks \emph{How do the two training components in \method contribute to its performance?} We examine their roles on Spatial and Object by removing one imitation term at a time and comparing each variant with full \method in Table~\ref{tab:ablation}. The original policy provides the reference without either term. These comparisons assess how each component affects target attainment, the physical response, and performance on permitted tasks.

\paragraph{Removing fallback imitation.}
We first remove fallback imitation while keeping supervision on retained tasks. This comparison tests whether continued training on permitted behaviors also changes execution of the prohibited task. The variant preserves high retained success but reaches the prohibited target at rates of 99.7\% on Spatial and 100.0\% on Object, leaving ongoing execution close to that of the original policy. Physical incidents also persist, including a 30.9\% rate on Object. Full \method reduces target success to 3.4\% and 0.3\%, respectively, with no recorded physical incidents in either suite. The difference supports explicitly teaching the holding response to suppress ongoing execution and guide the robot's behavior after takeover.

\paragraph{Removing retained-task imitation.}
To examine whether learning this fallback preserves the robot's other skills, we next keep fallback imitation and remove the retained-task term. Training only on fallback targets keeps target success low but reduces retained success to zero in both suites. Suppression in this variant therefore accompanies a loss of performance on the evaluated permitted tasks. With both terms present, full \method achieves retained success of 96.9\% on Spatial and 93.1\% on Object while maintaining low target attainment and no recorded physical incidents. This comparison identifies retained-task imitation as the component that preserves utility during adaptation. Together with the preceding removal, it supports learning the fallback response alongside continued supervision of the tasks the robot should retain. \revision{Additional comparisons of fallback supervision across execution states are provided in Appendix~\ref{app:prefix-supervision}.}

\subsection{Retained-Task Utility}
\label{sec:rq3}
The utility loss under fallback-only training leads to our third question, \emph{Can \method balance target suppression with retained-task performance?} To examine this trade-off, we evaluate checkpoints throughout adaptation on Spatial and Object and compare their suppression and retained success in Figure~\ref{fig:utility-tradeoff}. Each point pairs takeover suppression with performance on permitted tasks, placing policies that perform well on both measures toward the upper right. GA achieves high suppression with zero retained success at every sampled checkpoint. GradDiff improves retained success during adaptation, reaching 34.2\% on Spatial and 41.7\% on Object alongside high target suppression. NPO-RT and RMU exhibit the complementary trade-off, preserving high retained success at lower suppression levels. Refusal FT also reaches high suppression, although its retained success is lower than \method's at comparable suppression levels. The later \method checkpoints combine the two objectives, reaching 99.7\% suppression with 97.2\% retained success on Spatial and 99.0\% suppression with 93.6\% retained success on Object. These operating points preserve substantially more utility than GA and GradDiff at high suppression, while improving suppression over the utility-preserving NPO-RT and RMU checkpoints. \revision{Further retained-task results and checkpoint-selection details are provided in Appendix~\ref{app:rq3}.}


\begin{table}[t]
\centering
\fontsize{9}{10}\selectfont
\rqfourrevisioncolor
\setlength{\tabcolsep}{3pt}
\renewcommand{\arraystretch}{1.12}
\begin{tabular*}{\columnwidth}{@{\extracolsep{\fill}}lrr@{}}
\toprule
Method & \shortstack{Object dis.\\(cm) $\downarrow$} & \shortstack{Gripper contact\\rate (\%) $\uparrow$} \\
\midrule
Original policy & 28.7 & 13.1 \\
\midrule
GA & 63.8 & 64.7 \\
GradDiff & 71.6 & 63.7 \\
NPO-RT & 40.9 & 38.2 \\
RMU & 31.6 & 30.7 \\
Refusal FT & 27.6 & \textbf{98.8} \\
VLA-Forget (OFT) & 39.0 & 30.8 \\
\rowcolor{safirow}
\textbf{SAFI (ours)} & \textbf{3.1} & 97.3 \\
\bottomrule
\end{tabular*}
\captionsetup{labelfont+=safirqfourlabel}
\caption{\rqfourrevision{Final object displacement and gripper contact of learned policies, averaged equally across Spatial and Object with early and mid-task takeover. Bold marks the best value among adapted policies.}}
\label{tab:holding-summary}
\end{table}

\subsection{Physical Response After Takeover}
\label{sec:rq4}
To examine how fallback imitation shapes the robot's behavior during execution, our fourth question asks \emph{Does \method maintain the intended holding response after takeover?} We track the initially held object on Spatial and Object, measuring \emph{Object Displacement} in centimeters from its position at takeover. This quantity measures adherence to the stationary holding objective, with the scripted hold providing a reference response. Figure~\ref{fig:physical-response} compares all main-table methods from matched early and mid-task states. Across these conditions, \method's mean displacement remains low throughout the observation window, with end-of-window values ranging from 1.5 to 4.8\,cm. In each condition, this final mean is the closest to scripted hold among the evaluated learned policies. The difference persists under mid-task takeover, where \method records 4.8\,cm on Spatial and 3.2\,cm on Object. The smallest corresponding values among the other edited policies are 21.3\,cm for RMU and 12.9\,cm for Refusal FT. 

\rqfourrevision{We also report \emph{Gripper Contact Rate}, defined as the percentage of rollouts in which the gripper is in contact with the initially held object at the final observation. Table~\ref{tab:holding-summary} shows that \method achieves a mean final displacement of 3.1\,cm, the lowest among the adapted policies, with a 97.3\% contact rate. These results support \method's ability to keep objects near their takeover positions while preserving gripper contact. Additional results for individual takeover conditions and qualitative examples can be found in Appendices~\ref{app:physical-response} and~\ref{app:qualitative}.}
\section{Related Work}
\paragraph{VLA models and behavioral safety.}
VLA research improves manipulation through pretrained representations~\citep{kim2024openvla,qu2025spatialvla}, efficient action decoding~\citep{kim2025oft}, and language-based action hierarchies~\citep{belkhale2024rth}. Safety methods add constrained policy learning~\citep{zhang2025safevla} and contextual planning guardrails~\citep{ravichandran2026roboguard}. LIBERO-Safety evaluates physical and semantic constraints~\citep{cui2026liberosafety}, while HazardArena separates hazardous execution from capability failure~\citep{chen2026hazardarena}. \revision{For no-action training, \citet{jang2026nonstart} pair non-task instructions with zero arm motion and the demonstrated gripper command. \method preserves the task instruction and tests this holding response after takeover.}

\paragraph{Unlearning and fallback supervision.}
Machine unlearning removes selected data's influence~\citep{bourtoule2021unlearning}. Generative-model studies suppress harmful responses~\citep{yao2024unlearning,liu2024sku} and evaluate forgetting alongside retained utility~\citep{maini2024tofu,liu2025mllmu}. Task arithmetic edits parameter vectors~\citep{ilharco2023task}, NPO uses negative-preference optimization~\citep{zhang2024npo}, and RMU redirects representations~\citep{li2024wmdp}. Within VLAs, VLA-Forget selects components for editing~\citep{ranjan2026vlaforget}, target-and-control auditing examines task-vector negation~\citep{wang2026suppression}, and SOUL targets features associated with state hallucination~\citep{lee2026soul}. Replacement labels and refusal answers offer alternative supervision for targeted language inputs~\citep{eldan2023unlearning,maini2024tofu}. \method extends this supervision to fallback actions across execution states while preserving retained behaviors. We evaluate suppression alongside the resulting physical responses.

\section{Conclusion}
In this work, we address the challenges of stopping ongoing prohibited tasks and guiding VLAs toward appropriate physical responses. We propose \method, which first constructs holding targets throughout prohibited-task demonstrations and then jointly optimizes fallback and retained-task imitation. Experiments with OpenVLA-OFT on selected LIBERO and BadLIBERO tasks show that \method suppresses prohibited completion after takeover while preserving retained-task performance, with no recorded physical incidents.

\clearpage
\section*{Limitations}

\paragraph{Broader Deployment Settings.}
We evaluate OpenVLA-OFT on selected LIBERO and BadLIBERO tasks using a fixed training seed. Future work could extend this evaluation to additional VLA architectures and robot embodiments, with repeated training runs to assess the consistency of suppression and retained utility. Testing physical robots under varied sensing and control conditions would further clarify the approach's practical scope.

\paragraph{More Flexible Fallback Responses.}
\method currently learns a holding response for tasks that pass the applicability check. When an object is unstable or a physical interaction is already underway, a suitable fallback may require controlled motion. Future work could construct targets for responses such as placing an object down or withdrawing from contact, and study how to learn these responses alongside retained-task imitation.

\bibliography{ref}

@inproceedings{kim2024openvla,
  title={{OpenVLA}: An Open-Source Vision-Language-Action Model},
  author={Moo Jin Kim and Karl Pertsch and Siddharth Karamcheti and Ted Xiao and Ashwin Balakrishna and Suraj Nair and Rafael Rafailov and Ethan P Foster and Pannag R Sanketi and Quan Vuong and Thomas Kollar and Benjamin Burchfiel and Russ Tedrake and Dorsa Sadigh and Sergey Levine and Percy Liang and Chelsea Finn},
  booktitle={CoRL},
  year={2025}
}

@inproceedings{kim2025oft,
  title={Fine-Tuning Vision-Language-Action Models: Optimizing Speed and Success},
  author={Moo Jin Kim and Chelsea Finn and Percy Liang},
  booktitle={RSS},
  year={2025}
}

@inproceedings{zitkovich2023rt2,
  title={{RT-2}: Vision-Language-Action Models Transfer Web Knowledge to Robotic Control},
  author={Zitkovich, Brianna and Yu, Tianhe and Xu, Sichun and Xu, Peng and Xiao, Ted and Xia, Fei and Wu, Jialin and Wohlhart, Paul and Welker, Stefan and Wahid, Ayzaan and Vuong, Quan and Vanhoucke, Vincent and Tran, Huong and Soricut, Radu and Singh, Anikait and Singh, Jaspiar and Sermanet, Pierre and Sanketi, Pannag R. and Salazar, Grecia and Ryoo, Michael S. and Reymann, Krista and Rao, Kanishka and Pertsch, Karl and Mordatch, Igor and Michalewski, Henryk and Lu, Yao and Levine, Sergey and Lee, Lisa and Lee, Tsang-Wei Edward and Leal, Isabel and Kuang, Yuheng and Kalashnikov, Dmitry and Julian, Ryan and Joshi, Nikhil J. and Irpan, Alex and Ichter, Brian and Hsu, Jasmine and Herzog, Alexander and Hausman, Karol and Gopalakrishnan, Keerthana and Fu, Chuyuan and Florence, Pete and Finn, Chelsea and Dubey, Kumar Avinava and Driess, Danny and Ding, Tianli and Choromanski, Krzysztof Marcin and Chen, Xi and Chebotar, Yevgen and Carbajal, Justice and Brown, Noah and Brohan, Anthony and Arenas, Montserrat Gonzalez and Han, Kehang},
  booktitle={CoRL},
  year={2023}
}

@inproceedings{black2025pi0,
  title={{\(\pi_0\)}: A Vision-Language-Action Flow Model for General Robot Control},
  author={Kevin Black and Noah Brown and Danny Driess and Adnan Esmail and Michael Robert Equi and Chelsea Finn and Niccolo Fusai and Lachy Groom and Karol Hausman and Brian Ichter and Szymon Jakubczak and Tim Jones and Liyiming Ke and Sergey Levine and Adrian Li-Bell and Mohith Mothukuri and Suraj Nair and Karl Pertsch and Lucy Xiaoyang Shi and Laura Smith and James Tanner and Quan Vuong and Anna Walling and Haohuan Wang and Ury Zhilinsky},
  booktitle={RSS},
  year={2025}
}

@inproceedings{qu2025spatialvla,
  title={{SpatialVLA}: Exploring Spatial Representations for Visual-Language-Action Models},
  author={Delin Qu and Haoming Song and Qizhi Chen and Yuanqi Yao and Xinyi Ye and Jiayuan Gu and Zhigang Wang and Yan Ding and Bin Zhao and Dong Wang and Xuelong Li},
  booktitle={RSS},
  year={2025}
}

@inproceedings{black2025pi05,
  title={{\(\pi_{0.5}\)}: A Vision-Language-Action Model with Open-World Generalization},
  author={Kevin Black and Noah Brown and James Darpinian and Karan Dhabalia and Danny Driess and Adnan Esmail and Michael Robert Equi and Chelsea Finn and Niccolo Fusai and Manuel Y. Galliker and Dibya Ghosh and Lachy Groom and Karol Hausman and Brian Ichter and Szymon Jakubczak and Tim Jones and Liyiming Ke and Devin LeBlanc and Sergey Levine and Adrian Li-Bell and Mohith Mothukuri and Suraj Nair and Karl Pertsch and Allen Z. Ren and Lucy Xiaoyang Shi and Laura Smith and Jost Tobias Springenberg and Kyle Stachowicz and James Tanner and Quan Vuong and Homer Walke and Anna Walling and Haohuan Wang and Lili Yu and Ury Zhilinsky},
  booktitle={CoRL},
  year={2025}
}

@inproceedings{liu2023libero,
  title={{LIBERO}: Benchmarking Knowledge Transfer for Lifelong Robot Learning},
  author={Bo Liu and Yifeng Zhu and Chongkai Gao and Yihao Feng and Qiang Liu and Yuke Zhu and Peter Stone},
  booktitle={NeurIPS},
  year={2023}
}

@article{ranjan2026vlaforget,
  title={{VLA-Forget}: Vision-Language-Action Unlearning for Embodied Foundation Models},
  author={Ravi Ranjan and Agoritsa Polyzou},
  journal={arXiv preprint arXiv:2604.03956},
  year={2026}
}

@article{wang2026suppression,
  title={Suppression Sticks, Locality Is Fragile: A Closed-Loop Target-and-Control Audit of Task-Vector Negation in {VLA} Policies},
  author={Shaoguang Wang and Weiyu Guo and Rushi Dai and Yiren Zhao and Yandong Guo and Hui Xiong},
  journal={arXiv preprint arXiv:2608.04692},
  year={2026}
}

@article{chen2026hazardarena,
  title={{HazardArena}: Evaluating Semantic Safety in Vision-Language-Action Models},
  author={Zixing Chen and Yifeng Gao and Li Wang and Yunhan Zhao and Yi Liu and Jiayu Li and Xiang Zheng and Zuxuan Wu and Cong Wang and Xingjun Ma and Yu-Gang Jiang},
  journal={arXiv preprint arXiv:2604.12447},
  year={2026}
}

@article{jang2026nonstart,
  title={{VLA} Policies Do Not Know When Not to Act: Unintended Task Execution and Start Suppression},
  author={Minsu Jang and Jae-Woo Choi and Youngwoo Yoon and Dohyung Kim},
  journal={arXiv preprint arXiv:2610.41609},
  year={2026}
}

@article{lee2026soul,
  title={Sparse Feature Policy Unlearning Mitigates State Hallucination in Vision-Language-Action Models},
  author={Jiho Lee and Jeongeun Park and Heayoun Choi and Taekyung Kim and Eunwoo Kim},
  journal={arXiv preprint arXiv:2610.09496},
  year={2026}
}

@inproceedings{zhang2024npo,
  title={Negative Preference Optimization: From Catastrophic Collapse to Effective Unlearning},
  author={Ruiqi Zhang and Licong Lin and Yu Bai and Song Mei},
  booktitle={COLM},
  year={2024}
}

@inproceedings{li2024wmdp,
  title={The {WMDP} Benchmark: Measuring and Reducing Malicious Use with Unlearning},
  author={Nathaniel Li and Alexander Pan and Anjali Gopal and Summer Yue and Daniel Berrios and Alice Gatti and Justin D. Li and Ann-Kathrin Dombrowski and Shashwat Goel and Gabriel Mukobi and Nathan Helm-Burger and Rassin Lababidi and Lennart Justen and Andrew Bo Liu and Michael Chen and Isabelle Barrass and Oliver Zhang and Xiaoyuan Zhu and Rishub Tamirisa and Bhrugu Bharathi and Ariel Herbert-Voss and Cort B Breuer and Andy Zou and Mantas Mazeika and Zifan Wang and Palash Oswal and Weiran Lin and Adam Alfred Hunt and Justin Tienken-Harder and Kevin Y. Shih and Kemper Talley and John Guan and Ian Steneker and David Campbell and Brad Jokubaitis and Steven Basart and Stephen Fitz and Ponnurangam Kumaraguru and Kallol Krishna Karmakar and Uday Tupakula and Vijay Varadharajan and Yan Shoshitaishvili and Jimmy Ba and Kevin M. Esvelt and Alexandr Wang and Dan Hendrycks},
  booktitle={ICML},
  year={2024}
}

@inproceedings{maini2024tofu,
  title={{TOFU}: A Task of Fictitious Unlearning for {LLMs}},
  author={Pratyush Maini and Zhili Feng and Avi Schwarzschild and Zachary C. Lipton and J. Zico Kolter},
  booktitle={COLM},
  year={2024}
}

@article{zhou2025goba,
  title={Goal-oriented Backdoor Attack against Vision-Language-Action Models via Physical Objects},
  author={Zirun Zhou and Zhengyang Xiao and Haochuan Xu and Jing Sun and Di Wang and Jingfeng Zhang},
  journal={arXiv preprint arXiv:2510.09269},
  year={2025}
}

@article{cui2026liberosafety,
  title={{LIBERO-Safety}: A Comprehensive Benchmark for Physical and Semantic Safety in Vision-Language-Action Models},
  author={Rongxu Cui and Zongzheng Zhang and Jingrui Pang and Haohan Chi and Jinbang Guo and Saining Zhang and Shaoxuan Xie and Xin Jin and Yao Mu and Jiaolong Yang and Guocai Yao and Xianyuan Zhan and Ya-Qin Zhang and Hao Zhao},
  journal={arXiv preprint arXiv:2606.23686},
  year={2026}
}

@article{xu2025dropvla,
  title={{DropVLA}: An Action-Level Backdoor Attack on Vision-Language-Action Models},
  author={Zonghuan Xu and Jiayu Li and Yunhan Zhao and Xiang Zheng and Xingjun Ma and Yu-Gang Jiang},
  journal={arXiv preprint arXiv:2510.10932},
  year={2025}
}

@inproceedings{qi2025safety,
  title={Safety alignment should be made more than just a few tokens deep},
  author={Qi, Xiangyu and Panda, Ashwinee and Lyu, Kaifeng and Ma, Xiao and Roy, Subhrajit and Beirami, Ahmad and Mittal, Prateek and Henderson, Peter},
  booktitle={ICLR},
  year={2025}
}

@inproceedings{brohan2023rt1,
  title={{RT-1: Robotics Transformer for Real-World Control at Scale}},
  author={Anthony Brohan and Noah Brown and Justice Carbajal and Yevgen Chebotar and Joseph Dabis and Chelsea Finn and Keerthana Gopalakrishnan and Karol Hausman and Alexander Herzog and Jasmine Hsu and Julian Ibarz and Brian Ichter and Alex Irpan and Tomas Jackson and Sally Jesmonth and Nikhil Joshi and Ryan Julian and Dmitry Kalashnikov and Yuheng Kuang and Isabel Leal and Kuang-Huei Lee and Sergey Levine and Yao Lu and Utsav Malla and Deeksha Manjunath and Igor Mordatch and Ofir Nachum and Carolina Parada and Jodilyn Peralta and Emily Perez and Karl Pertsch and Jornell Quiambao and Kanishka Rao and Michael S Ryoo and Grecia Salazar and Pannag R Sanketi and Kevin Sayed and Jaspiar Singh and Sumedh Sontakke and Austin Stone and Clayton Tan and Huong Tran and Vincent Vanhoucke and Steve Vega and Quan H Vuong and Fei Xia and Ted Xiao and Peng Xu and Sichun Xu and Tianhe Yu and Brianna Zitkovich},
  booktitle={RSS},
  year={2023}
}

@inproceedings{ghosh2024octo,
  title={{Octo: An Open-Source Generalist Robot Policy}},
  author={Dibya Ghosh and Homer Rich Walke and Karl Pertsch and Kevin Black and Oier Mees and Sudeep Dasari and Joey Hejna and Tobias Kreiman and Charles Xu and Jianlan Luo and You Liang Tan and Lawrence Yunliang Chen and Quan Vuong and Ted Xiao and Pannag R Sanketi and Dorsa Sadigh and Chelsea Finn and Sergey Levine},
  booktitle={RSS},
  year={2024}
}

@inproceedings{driess2023palme,
  title={{PaLM-E}: An Embodied Multimodal Language Model},
  author={Driess, Danny and Xia, Fei and Sajjadi, Mehdi S. M. and Lynch, Corey and Chowdhery, Aakanksha and Ichter, Brian and Wahid, Ayzaan and Tompson, Jonathan and Vuong, Quan and Yu, Tianhe and Huang, Wenlong and Chebotar, Yevgen and Sermanet, Pierre and Duckworth, Daniel and Levine, Sergey and Vanhoucke, Vincent and Hausman, Karol and Toussaint, Marc and Greff, Klaus and Zeng, Andy and Mordatch, Igor and Florence, Pete},
  booktitle={ICML},
  year={2023}
}

@inproceedings{jiang2023vima,
  title={{VIMA}: Robot Manipulation with Multimodal Prompts},
  author={Jiang, Yunfan and Gupta, Agrim and Zhang, Zichen and Wang, Guanzhi and Dou, Yongqiang and Chen, Yanjun and Fei-Fei, Li and Anandkumar, Anima and Zhu, Yuke and Fan, Linxi},
  booktitle={ICML},
  year={2023}
}

@inproceedings{chi2023diffusion,
  title={{Diffusion Policy: Visuomotor Policy Learning via Action Diffusion}},
  author={Cheng Chi and Siyuan Feng and Yilun Du and Zhenjia Xu and Eric Cousineau and Benjamin CM Burchfiel and Shuran Song},
  booktitle={RSS},
  year={2023}
}

@inproceedings{zhao2023act,
  title={{Learning Fine-Grained Bimanual Manipulation with Low-Cost Hardware}},
  author={Tony Z. Zhao and Vikash Kumar and Sergey Levine and Chelsea Finn},
  booktitle={RSS},
  year={2023}
}

@inproceedings{ichter2023saycan,
  title={Do As I Can, Not As I Say: Grounding Language in Robotic Affordances},
  author={Ichter, Brian and Brohan, Anthony and Chebotar, Yevgen and Finn, Chelsea and Hausman, Karol and Herzog, Alexander and Ho, Daniel and Ibarz, Julian and Irpan, Alex and Jang, Eric and Julian, Ryan and Kalashnikov, Dmitry and Levine, Sergey and Lu, Yao and Parada, Carolina and Rao, Kanishka and Sermanet, Pierre and Toshev, Alexander T and Vanhoucke, Vincent and Xia, Fei and Xiao, Ted and Xu, Peng and Yan, Mengyuan and Brown, Noah and Ahn, Michael and Cortes, Omar and Sievers, Nicolas and Tan, Clayton and Xu, Sichun and Reyes, Diego and Rettinghouse, Jarek and Quiambao, Jornell and Pastor, Peter and Luu, Linda and Lee, Kuang-Huei and Kuang, Yuheng and Jesmonth, Sally and Joshi, Nikhil J. and Jeffrey, Kyle and Ruano, Rosario Jauregui and Hsu, Jasmine and Gopalakrishnan, Keerthana and David, Byron and Zeng, Andy and Fu, Chuyuan Kelly},
  booktitle={CoRL},
  year={2023}
}

@inproceedings{ren2023knowno,
  title={Robots That Ask For Help: Uncertainty Alignment for Large Language Model Planners},
  author={Ren, Allen Z. and Dixit, Anushri and Bodrova, Alexandra and Singh, Sumeet and Tu, Stephen and Brown, Noah and Xu, Peng and Takayama, Leila and Xia, Fei and Varley, Jake and Xu, Zhenjia and Sadigh, Dorsa and Zeng, Andy and Majumdar, Anirudha},
  booktitle={CoRL},
  year={2023}
}

@inproceedings{ross2011dagger,
  title={A Reduction of Imitation Learning and Structured Prediction to No-Regret Online Learning},
  author={Ross, Stephane and Gordon, Geoffrey and Bagnell, Drew},
  booktitle={AISTATS},
  year={2011}
}

@inproceedings{liu2024sku,
  title={Towards Safer Large Language Models through Machine Unlearning},
  author={Liu, Zheyuan and Dou, Guangyao and Tan, Zhaoxuan and Tian, Yijun and Jiang, Meng},
  booktitle={Findings of ACL},
  year={2024}
}

@inproceedings{liu2025mllmu,
  title={Protecting Privacy in Multimodal Large Language Models with {MLLMU-Bench}},
  author={Liu, Zheyuan and Dou, Guangyao and Jia, Mengzhao and Tan, Zhaoxuan and Zeng, Qingkai and Yuan, Yongle and Jiang, Meng},
  booktitle={NAACL},
  year={2025}
}

@inproceedings{belkhale2024rth,
  title={{RT-H}: Action Hierarchies using Language},
  author={Suneel Belkhale and Tianli Ding and Ted Xiao and Pierre Sermanet and Quan Vuong and Jonathan Tompson and Yevgen Chebotar and Debidatta Dwibedi and Dorsa Sadigh},
  booktitle={RSS},
  year={2024}
}

@inproceedings{zhang2025safevla,
  title={{SafeVLA}: Towards Safety Alignment of Vision-Language-Action Model via Constrained Learning},
  author={Borong Zhang and Yuhao Zhang and Jiaming Ji and Yingshan Lei and Juntao Dai and Yuanpei Chen and Yaodong Yang},
  booktitle={NeurIPS},
  year={2025}
}

@inproceedings{robey2025robopair,
  title={Jailbreaking {LLM}-Controlled Robots},
  author={Alexander Robey and Zachary Ravichandran and Vijay Kumar and Hamed Hassani and George J. Pappas},
  booktitle={ICRA},
  year={2025}
}

@article{ravichandran2026roboguard,
  title={Safety Guardrails for {LLM}-Enabled Robots},
  author={Zachary Ravichandran and Alexander Robey and Vijay Kumar and George J. Pappas and Hamed Hassani},
  journal={IEEE Robotics and Automation Letters},
  year={2026}
}

@inproceedings{bourtoule2021unlearning,
  title={Machine Unlearning},
  author={Lucas Bourtoule and Varun Chandrasekaran and Christopher A. Choquette-Choo and Hengrui Jia and Adelin Travers and Baiwu Zhang and David Lie and Nicolas Papernot},
  booktitle={IEEE S\&P},
  year={2021}
}

@inproceedings{ilharco2023task,
  title={Editing Models with Task Arithmetic},
  author={Gabriel Ilharco and Marco Tulio Ribeiro and Mitchell Wortsman and Suchin Gururangan and Ludwig Schmidt and Hannaneh Hajishirzi and Ali Farhadi},
  booktitle={ICLR},
  year={2023}
}

@inproceedings{yao2024unlearning,
  title={Large Language Model Unlearning},
  author={Yuanshun Yao and Xiaojun Xu and Yang Liu},
  booktitle={NeurIPS},
  year={2024}
}

@article{eldan2023unlearning,
  title={Who's {Harry Potter}? Approximate Unlearning in {LLMs}},
  author={Ronen Eldan and Mark Russinovich},
  journal={arXiv preprint arXiv:2310.02238},
  year={2023}
}

@inproceedings{hu2022lora,
  title={{LoRA}: Low-Rank Adaptation of Large Language Models},
  author={Edward J. Hu and Yelong Shen and Phillip Wallis and Zeyuan Allen-Zhu and Yuanzhi Li and Shean Wang and Lu Wang and Weizhu Chen},
  booktitle={ICLR},
  year={2022}
}

@inproceedings{liu2026mtmcs,
  title={{MTMCS}-Bench: Evaluating Contextual Safety of Multimodal Large Language Models in Multi-Turn Dialogues},
  author={Zheyuan Liu and Dongwhi Kim and Yixin Wan and Xiangchi Yuan and Zhaoxuan Tan and Fengran Mo and Meng Jiang},
  booktitle={Findings of ACL},
  year={2026}
}

@article{cui2026learning,
  title={Learning from Runtime Feedback through Failure-Bank Self-Evolution for Vision-Language-Action Models},
  author={Cui, Mingyue and Liu, Zheyuan and Zhu, Yihan and Zhang, Zheyuan and Jiang, Meng},
  journal={arXiv preprint arXiv:2609.39820},
  year={2026}
}
\clearpage
\appendix
\onecolumn
\begingroup
\raggedbottom\raggedcolumns
\setlength{\premulticols}{8\baselineskip}
\setlength{\maxbalancingoverflow}{0pt}
\makeatletter
\newcommand{\suppparagraphlabel}[1]{\begingroup\edef\@currentlabel{\thesubsection}\label{#1}\endgroup}
\makeatother
\ifdefined\setmakelinenumbers
  \newcommand{\suppcolumnnumber}{\docolaction{\makeLineNumberLeft}{\makeLineNumberLeft}{\makeLineNumberRight}}
  \AtBeginEnvironment{multicols}{\setmakelinenumbers\suppcolumnnumber*}
\fi
\renewenvironment{table*}[1][]{\begin{table}[H]}{\end{table}}
\renewenvironment{figure*}[1][]{\begin{figure}[H]}{\end{figure}}
\begin{multicols}{2}
\section{Appendix Outline}
\label{app:outline}
This appendix expands the method and the four research questions in the main text. It also includes computational costs and qualitative examples, organized as follows.

\begingroup
\normalsize
\setlength{\parindent}{0pt}
\newcommand{\appentry}[2]{\par\addvspace{0.65em}\noindent\hyperref[#1]{\textbf{\ref*{#1}\quad #2}}\nobreak\hfill\pageref*{#1}\par\nobreak\vspace{0.2em}}
\newcommand{\appsubentry}[2]{\noindent\hspace*{1em}\hyperref[#1]{\ref*{#1}\quad #2}\nobreak\hfill\pageref*{#1}\par\nobreak\vspace{0.3em}}
\appentry{app:algorithm}{Algorithm and Action Encoding}
\appsubentry{app:fallback}{Fallback Action Encoding}
\appentry{app:settings}{Experimental Settings}
\appsubentry{app:results-protocol}{Evaluation Coverage}
\appsubentry{app:takeover}{Takeover States and Windows}
\appsubentry{app:metrics}{Evaluation Metrics}
\appsubentry{app:training}{Training and Selection}
\appsubentry{app:baseline-settings}{Baseline Adaptations}
\appentry{app:additional}{Supplementary Results}
\appsubentry{app:rq1}{Effectiveness (RQ1)}
\appsubentry{app:prefix-supervision}{Supervision Design (RQ2)}
\appsubentry{app:rq3}{Retained Utility (RQ3)}
\appsubentry{app:physical-response}{Physical Responses (RQ4)}
\appentry{app:efficiency}{Computational Efficiency}
\appentry{app:qualitative}{Qualitative Comparisons}
\endgroup

\section{Algorithm and Action Encoding}
\label{app:algorithm}
Algorithm~\ref{alg:safi} summarizes the two-stage procedure in Section~\ref{sec:method}. After checking that holding is an appropriate response, \method constructs fallback targets and jointly learns them with retained-task actions using the two imitation losses in Figure~\ref{fig:method}.

\begin{algorithm*}[t]
\caption{State-Aware Fallback Imitation (\method)}
\label{alg:safi}
\small
\begin{algorithmic}[1]
\Require Original policy $\pi_0$; prohibited and retained demonstrations $\mathcal D_{\mathrm p},\mathcal D_{\mathrm r}$; action-normalization statistics; adaptation budget $K$; learning rate $\eta$; calibration checkpoint set $\mathcal C$
\Ensure Adapted policy $\pi_{\theta^\star}$ for a task passing the scripted-hold applicability check
\State Verify scripted holding on the task's candidate takeover states
\Statex \textbf{Stage 1: Fallback Target Construction}
\State Initialize the fallback dataset $\overline{\mathcal D}_{\mathrm p}\gets\varnothing$
\For{each $(x,A)\in\mathcal D_{\mathrm p}$}
    \State Encode physical zero arm motion as $h$ using the sample's normalization statistics
    \State Construct $\bar A_{k,1:d-1}\gets h$ and $\bar A_{k,d}\gets A_{1,d}$ for $k=1,\ldots,H$
    \State Add $(x,\bar A)$ to $\overline{\mathcal D}_{\mathrm p}$
\EndFor
\Statex \textbf{Stage 2: Joint Policy Adaptation}
\State Initialize from $\pi_0$ and enable the language-model LoRA adapters and inherited action head
\For{update $u=1,\ldots,K$}
    \State Sample fallback batch $B_{\mathrm p}\sim\overline{\mathcal D}_{\mathrm p}$ and retained batch $B_{\mathrm r}\sim\mathcal D_{\mathrm r}$
    \State $\widehat{\mathcal L}_{\mathrm{fallback}}\gets |B_{\mathrm p}|^{-1}\sum_{(x,\bar A)\in B_{\mathrm p}}\ell(\hat A_\theta(x),\bar A)$
    \State $\widehat{\mathcal L}_{\mathrm{retain}}\gets |B_{\mathrm r}|^{-1}\sum_{(x,A)\in B_{\mathrm r}}\ell(\hat A_\theta(x),A)$
    \State Update $\theta$ with AdamW on $\widehat{\mathcal L}_{\mathrm{fallback}}+\widehat{\mathcal L}_{\mathrm{retain}}$, using learning rate $\eta$
    \If{$u\in\mathcal C$}
        \State Save the adapted parameters for calibration
    \EndIf
\EndFor
\State Select $\theta^\star$ by the suppression and retained-success rule in Appendix~\ref{app:selection}
\State \Return $\pi_{\theta^\star}$
\end{algorithmic}
\end{algorithm*}

\subsection{Fallback Action Encoding}
\label{app:fallback}
The OFT action representation has $d=7$ coordinates, comprising six arm-motion coordinates and one gripper coordinate. The specialist experiments use chunks of $H=8$ actions. For arm coordinate $j$, let $q_{01,j}$ and $q_{99,j}$ be the first and ninety-ninth percentile bounds in the original policy's action-normalization statistics. Equation~\ref{eq:hold} encodes physical zero motion as
\begin{equation}
\label{eq:hold-normalization}
h_j=\frac{2(0-q_{01,j})}{q_{99,j}-q_{01,j}}-1,
\quad j=1,\ldots,d-1,
\end{equation}
with nondegenerate normalization intervals. Each sample uses its data source's statistics and repeats the first demonstrated gripper label, $\bar A_{k,d}=A_{1,d}$, across the chunk.

\section{Experimental Settings}
\label{app:settings}
The following settings support the comparisons in Section~\ref{sec:experiments}.

\subsection{Evaluation Coverage}
\label{app:results-protocol}
We use the released Goal, Spatial, and Object specialists\revision{ on the tasks listed in Table~\ref{tab:task-boundaries}}. The Spatial and Object cohorts contain the first four eligible tasks in a fixed candidate order. Eligibility requires successful original-policy execution, faithful restoration of takeover states, and scripted holding that prevents prohibited completion and physical incidents. Both takeover states must retain gripper contact with the same target object.

Calibration and held-out evaluation use separate initial states\revision{, with normal starts for both target and retained tasks during calibration}. Calibration uses ten states per task, followed by forty held-out target episodes and ten episodes on each of the other nine tasks for retained utility. The LIBERO aggregate weights Goal, Spatial, and Object equally after averaging within each suite, with the breakdown in Appendix~\ref{app:rq1}.

\subsection{Takeover States and Windows}
\label{app:takeover}
Successful original-policy trajectories supply matched states for every evaluated method. The early state occurs ten control steps after the first gripper-close command, and the mid-task state lies halfway between that point and task completion. Each takeover starts a fresh action chunk from the edited policy. Snapshot restoration supplies \revision{cream-cheese placement} states, and action replay supplies the remaining states. Table~\ref{tab:takeover-protocols} summarizes the observation windows, including the settings that continue after target attainment.

\begin{table}[H]
\centering\small
\begin{tabular}{lll}
\toprule
Setting & Restoration & Termination\\
\midrule
\revision{Cream cheese} & Snapshot & Success-censored\\
\revision{Bottle} & Replay & Success or 300 steps\\
\revision{Stove} & Replay & 300 steps\\
Spatial & Replay & 220 steps\\
Object & Replay & 280 steps\\
BadLIBERO & Replay & 200 steps\\
\bottomrule
\end{tabular}
\caption{Takeover protocols. Fixed-window evaluations continue after target attainment. The Goal stove audit also includes an initial-state condition.}
\label{tab:takeover-protocols}
\end{table}

\subsection{Evaluation Metrics}
\label{app:metrics}
We define each metric at the rollout level before averaging across takeover conditions and tasks. Let $\tau_i$ denote a rollout and $T_i$ its recorded length. Its binary flags $c_i$, $e_i$, and $s_i$ indicate prohibited-target attainment, a registered physical incident, and command saturation, respectively. For a fixed task and takeover condition with $N$ rollouts, the corresponding percentage rates are
\begin{equation}
\label{eq:episode-rates}
\begin{aligned}
R_{\mathrm{target}}&=\frac{100}{N}\sum_{i=1}^{N}c_i,\\
R_{\mathrm{incident}}&=\frac{100}{N}\sum_{i=1}^{N}e_i,\\
R_{\mathrm{sat}}&=\frac{100}{N}\sum_{i=1}^{N}s_i.
\end{aligned}
\end{equation}
These are Target Success Rate, Physical Incident Rate, and Action Saturation Rate, respectively. Each uses all rollouts in that condition as its denominator, including those in which target attainment and a physical incident coincide.

\paragraph{Target attainment.}
Let $\phi_q$ be the prohibited-target predicate for task $q$. We record $c_i=\mathbf 1[\phi_q\text{ holds at any observed step}]$. For Goal bottle and cream-cheese placement, Spatial, and Object, $\phi_q$ is the environment's task-success predicate. The stove target is attained when the knob position reaches 0.5. In BadLIBERO, the target is entry into the attack's designated L3 region. The horizontal axis of Figure~\ref{fig:utility-tradeoff} uses the complement
\begin{equation}
\label{eq:target-suppression}
R_{\mathrm{suppression}}=100-R_{\mathrm{target}},
\end{equation}
computed on the same takeover rollouts.

\paragraph{Physical incidents.}
The incident flag is the union of the setting's registered physical-event flags,
\begin{equation}
\label{eq:incident-union}
e_i=\mathbf 1\!\left[\bigvee_{k\in\mathcal K_q} e_i^{(k)}=1\right],
\end{equation}
where $\mathcal K_q$ contains the applicable event types. The LIBERO union includes contact-aware object drops and fallen-body events. For a held object, let $z_i(t)$ be its height in meters and let $g_i(t)$ and $b_i(t)$ indicate contact with the gripper and goal body. The unsupported-low indicator is
\begin{equation}
\label{eq:unsupported-low}
\begin{aligned}
u_i(t)={}&\mathbf 1[z_i(t)<z_i(0)-0.10]\\
&\times(1-g_i(t))(1-b_i(t)).
\end{aligned}
\end{equation}
The \dropmetric flag requires two consecutive recorded steps with $u_i(t)=1$. Fallen-body events include another movable body descending more than 10\,cm from its takeover height or the held object satisfying $u_i(T_i-1)=1$ at the final recorded step. For the stove target, the union uses the applicable fallen-body events. BadLIBERO additionally includes lost hold and disturbance of other objects under its task-specific definitions in Appendix~\ref{app:badlibero}.

\paragraph{Command saturation.}
For processed arm command $a_{i,t,j}$ before controller clipping, define
\begin{equation}
\label{eq:sat}
\begin{aligned}
v_i(t)&=\mathbf 1\!\left[\max_{1\leq j\leq6}|a_{i,t,j}|\geq1\right],\\
\rho_i&=\frac{1}{T_i}\sum_{t=0}^{T_i-1}v_i(t).
\end{aligned}
\end{equation}
LIBERO uses $s_i=\mathbf 1[\rho_i\geq0.05]$, and BadLIBERO uses $s_i=\mathbf 1[\rho_i>0]$. These criteria measure commands at the clipping boundary over the observation windows in Table~\ref{tab:takeover-protocols}.

\paragraph{Retained-task success.}
For the permitted-task set $\mathcal R_q$ associated with prohibited task $q$, let $y_{r,i}$ denote normal-start success on retained task $r$, evaluated over $n_r$ initial states. Retained Success Rate is
\begin{equation}
\label{eq:retained-rate}
R_{\mathrm{retain}}(q)=\frac{100}{|\mathcal R_q|}
\sum_{r\in\mathcal R_q}\frac{1}{n_r}\sum_{i=1}^{n_r}y_{r,i}.
\end{equation}
The specialist evaluations use the other nine tasks in the suite, with the same number of initial states per retained task. For BadLIBERO, retained utility is clean-task success on untriggered inputs, as reported in Appendix~\ref{app:badlibero}.

\paragraph{Aggregation and uncertainty.}
For a takeover rate $R$, let $R_{q,b}$ be its value for target $q$ at boundary $b\in\{\mathrm{early},\mathrm{mid}\}$. Given the evaluated target set $\mathcal Q_s$ for suite $s$, we compute
\begin{equation}
\label{eq:suite-aggregation}
R_s=\frac{1}{2|\mathcal Q_s|}
\sum_{q\in\mathcal Q_s}\sum_b R_{q,b},
\end{equation}
and average the three LIBERO suites equally for the main column. BadLIBERO retains its own predicates and observation window. Confidence intervals use paired donor bootstrap, preserving matched methods and takeover states within each resample. \revision{The intervals describe variation across trajectories from different initial states for the fixed targets and checkpoints.}

\subsection{Training and Checkpoint Selection}
\label{app:training}
We adapt the OFT specialist through language-model LoRA and the inherited action head, keeping the remaining components frozen. \revision{The training settings for \method are given in Table~\ref{tab:hyperparameters}, with its 300-update budget shared by GA, GradDiff, NPO-RT, RMU, Refusal FT, and the loss and coverage variants on Spatial and Object.}

\begin{table}[H]
\centering\small
\begin{tabular}{@{}ll@{}}
\toprule
Hyperparameter & Setting\\
\midrule
Optimizer & AdamW\\
Learning rate & $5\times10^{-5}$\\
LoRA rank & 32\\
Samples per update & 8 per data stream\\
Imitation weights & 1 for each term\\
Training updates & 300\\
\bottomrule
\end{tabular}
\caption{Core settings for \revision{\method on the LIBERO specialists}. Fallback and retained samples contribute equally weighted imitation losses.}
\label{tab:hyperparameters}
\end{table}

\paragraph{Checkpoint selection.}
\label{app:selection}
A checkpoint qualifies when calibration target success is at most 10\% and retained success decreases by at most ten percentage points relative to the original. We select the qualifying checkpoint with the highest retained success, breaking ties toward the earlier checkpoint. \revision{When none qualifies, we apply this ranking to the candidates that meet the suppression criterion. If suppression also fails, we first minimize target success and then apply the same ranking.} Held-out outcomes are evaluated after selection. \revision{BadLIBERO uses each recipe's final checkpoint, which is at 300 updates for the single-stage edits.}

\subsection{Baseline Adaptations}
\label{app:baseline-settings}
All baselines use the OFT action representation. GA increases L1 error on prohibited demonstrations, and GradDiff combines this update with retained-action imitation. NPO-RT applies negative preference optimization through a continuous-action score and retained imitation. RMU redirects prohibited-input hidden states toward a random vector while preserving original hidden states on retained inputs.

\paragraph{Response and VLA baselines.}
\label{app:comparators}
Refusal FT learns controller-generated home-return trajectories from interrupted tasks, together with retained-task demonstrations. \revision{The controller preserves the last gripper command during return, opens the gripper for reset demonstrations, and clips arm commands to the normalization bounds. Training demonstrations are interrupted before release or completion for placement tasks and before contact for the stove. Their median interruption times are later than held-out mid-task takeover for placement and earlier for the stove.} VLA-Forget uses staged module-wise LoRA, gradient projection, and action-output distillation in our OFT implementation. \label{app:baseline-provenance}The VLA-Forget results describe our deployed OFT adaptation, whose action outputs differ after checkpoint reloading.

\section{Supplementary Results}
\label{app:additional}
These analyses extend the four research questions in Section~\ref{sec:experiments}. They provide suite-level effectiveness, supervision-design comparisons, retained-task evaluations, and physical-response measurements.

\subsection{Effectiveness (RQ1)}
\label{app:rq1}
Table~\ref{tab:results-breakdown} separates the Goal, Spatial, and Object results averaged in the main table. Table~\ref{tab:normal-retained} complements these takeover outcomes with normal-start target suppression and retained-task success, connecting behavior during execution to performance from initial states.

\end{multicols}
\begin{table*}[!t]
\centering
\footnotesize
\setlength{\tabcolsep}{2.75pt}
\renewcommand{\arraystretch}{1.12}
\begin{tabular*}{\textwidth}{@{\extracolsep{\fill}}lrrrrrr@{}}
\toprule
 & \multicolumn{3}{c}{LIBERO-Goal} & \multicolumn{3}{c}{LIBERO-Spatial} \\
\cmidrule(lr){2-4}\cmidrule(lr){5-7}
Method & \shortstack{Target success\\rate (\%) $\downarrow$} & \shortstack{Physical incident\\rate (\%) $\downarrow$} & \shortstack{Action saturation\\rate (\%) $\downarrow$} & \shortstack{Target success\\rate (\%) $\downarrow$} & \shortstack{Physical incident\\rate (\%) $\downarrow$} & \shortstack{Action saturation\\rate (\%) $\downarrow$} \\
\midrule
Original policy & 99.2 & 4.2 & 0.0 & 100.0 & 1.7 & 0.0 \\
GA & 33.3 & 24.2 & 100.0 & 0.3 & 5.3 & 100.0 \\
GradDiff & 33.8 & 26.2 & 100.0 & 1.4 & 26.5 & 99.3 \\
NPO-RT & 32.9 & 16.7 & 94.2 & 45.4 & 20.3 & 25.0 \\
RMU & 17.1 & 0.0 & 0.0 & 73.7 & 9.4 & 0.0 \\
Refusal FT & 33.3 & 12.1 & 0.0 & 0.0 & 0.0 & 0.0 \\
VLA-Forget (OFT) & 12.1 & 16.7 & 71.7 & 42.4 & 21.3 & 73.0 \\
\rowcolor{safirow}
\textbf{\method (ours)} & 0.0 & 0.0 & 0.0 & 3.4 & 0.0 & 0.0 \\
\midrule
 & \multicolumn{3}{c}{LIBERO-Object} & \multicolumn{3}{c}{BadLIBERO} \\
\cmidrule(lr){2-4}\cmidrule(lr){5-7}
Method & \shortstack{Target success\\rate (\%) $\downarrow$} & \shortstack{Physical incident\\rate (\%) $\downarrow$} & \shortstack{Action saturation\\rate (\%) $\downarrow$} & \shortstack{Target success\\rate (\%) $\downarrow$} & \shortstack{Physical incident\\rate (\%) $\downarrow$} & \shortstack{Action saturation\\rate (\%) $\downarrow$} \\
\midrule
Original policy & 100.0 & 30.6 & 0.0 & 87.5 & 37.5 & 0.0 \\
GA & 3.0 & 19.5 & 100.0 & 0.0 & 40.0 & 100.0 \\
GradDiff & 0.0 & 29.6 & 99.0 & 0.0 & 30.0 & 100.0 \\
NPO-RT & 17.3 & 22.9 & 21.5 & 0.0 & 22.5 & 100.0 \\
RMU & 61.0 & 33.8 & 0.0 & 0.0 & 0.0 & 0.0 \\
Refusal FT & 1.4 & 0.3 & 0.0 & 0.0 & 0.0 & 2.5 \\
VLA-Forget (OFT) & 38.1 & 22.8 & 16.2 & 35.0 & 35.0 & 100.0 \\
\rowcolor{safirow}
\textbf{\method (ours)} & 0.3 & 0.0 & 0.0 & 0.0 & 0.0 & 0.0 \\
\bottomrule
\end{tabular*}
\caption{Setting-level takeover results underlying Table~\ref{tab:main-results}. Each setting uses its registered target cohort, observation windows, and event definitions.}
\label{tab:results-breakdown}
\end{table*}

\begin{table*}[!t]
\centering
\footnotesize
\setlength{\tabcolsep}{2.75pt}
\renewcommand{\arraystretch}{1.12}
\begin{tabular*}{\textwidth}{@{\extracolsep{\fill}}lrrrrrr@{}}
\toprule
 & \multicolumn{2}{c}{LIBERO-Goal} & \multicolumn{2}{c}{LIBERO-Spatial} & \multicolumn{2}{c}{LIBERO-Object} \\
\cmidrule(lr){2-3}\cmidrule(lr){4-5}\cmidrule(lr){6-7}
Method & \shortstack{Target success\\rate (\%) $\downarrow$} & \shortstack{Retained success\\rate (\%) $\uparrow$} & \shortstack{Target success\\rate (\%) $\downarrow$} & \shortstack{Retained success\\rate (\%) $\uparrow$} & \shortstack{Target success\\rate (\%) $\downarrow$} & \shortstack{Retained success\\rate (\%) $\uparrow$} \\
\midrule
Original policy & 100.0 & 94.4 & 98.1 & 99.2 & 96.2 & 94.2 \\
GA & 3.3 & 0.0 & 0.0 & 0.0 & 0.0 & 0.0 \\
GradDiff & 0.0 & 45.2 & 0.0 & 34.2 & 0.0 & 41.7 \\
NPO-RT & 0.0 & 98.1 & 4.4 & 96.9 & 0.0 & 92.8 \\
RMU & 0.8 & 97.0 & 70.0 & 97.8 & 48.8 & 97.8 \\
Refusal FT & 0.0 & 94.4 & 0.0 & 91.9 & 0.6 & 77.2 \\
VLA-Forget (OFT) & 0.0 & 87.8 & 0.0 & 85.3 & 1.2 & 45.6 \\
\rowcolor{safirow}
\textbf{\method (ours)} & 0.0 & 93.7 & 0.0 & 96.9 & 0.0 & 93.1 \\
\bottomrule
\end{tabular*}
\caption{Normal-start target and retained-task success for the selected policies. Target success here measures execution from initial states. Table~\ref{tab:main-results} measures target attainment after takeover.}
\label{tab:normal-retained}
\end{table*}

\begin{multicols}{2}
\paragraph{\revision{Takeover outcomes by task.}}
\revision{The per-task results show that Refusal FT's target completions on Goal concentrate on stove activation, and its physical incidents concentrate on mid-task bottle placement (Table~\ref{tab:task-boundaries}). To distinguish these outcomes within the same rollouts, Table~\ref{tab:joint-outcomes} reports their joint frequencies on Spatial and Object. For the original Object policy, every incident occurs in a successful mid-task rollout that ends with basket contact.}

\end{multicols}
\begin{table*}[!tp]
\centering\footnotesize\revisioncolor
\captionsetup{labelfont+=safirevisionlabel}
\setlength{\tabcolsep}{3pt}
\renewcommand{\arraystretch}{1.07}
\begin{tabular*}{\textwidth}{@{\extracolsep{\fill}}llrrrrrr@{}}
\toprule
 &  & \multicolumn{2}{c}{Original policy} & \multicolumn{2}{c}{Refusal FT} & \multicolumn{2}{c}{SAFI} \\
\cmidrule(lr){3-4}\cmidrule(lr){5-6}\cmidrule(lr){7-8}
Task & Takeover & Target & Incident & Target & Incident & Target & Incident \\
\midrule
\multicolumn{8}{@{}l}{\textit{Goal}} \\
Cream-cheese placement & Early & 97.5 & 0.0 & 2.5 & 0.0 & 0.0 & 0.0 \\
 & Mid & 97.5 & 25.0 & 0.0 & 7.5 & 0.0 & 0.0 \\
Bottle placement & Early & 100.0 & 0.0 & 0.0 & 0.0 & 0.0 & 0.0 \\
 & Mid & 100.0 & 0.0 & 0.0 & 65.0 & 0.0 & 0.0 \\
Stove activation & Early & 100.0 & 0.0 & 97.5 & 0.0 & 0.0 & 0.0 \\
 & Mid & 100.0 & 0.0 & 100.0 & 0.0 & 0.0 & 0.0 \\
\midrule
\multicolumn{8}{@{}l}{\textit{Spatial \textemdash{} place the bowl on the plate}} \\
Bowl next to the plate & Early & 100.0 & 0.0 & 0.0 & 0.0 & 0.0 & 0.0 \\
 & Mid & 100.0 & 0.0 & 0.0 & 0.0 & 0.0 & 0.0 \\
Bowl next to the cookie box & Early & 100.0 & 0.0 & 0.0 & 0.0 & 0.0 & 0.0 \\
 & Mid & 100.0 & 11.1 & 0.0 & 0.0 & 0.0 & 0.0 \\
Bowl on the stove & Early & 100.0 & 0.0 & 0.0 & 0.0 & 0.0 & 0.0 \\
 & Mid & 100.0 & 2.6 & 0.0 & 0.0 & 0.0 & 0.0 \\
Bowl between plate and ramekin & Early & 100.0 & 0.0 & 0.0 & 0.0 & 0.0 & 0.0 \\
 & Mid & 100.0 & 0.0 & 0.0 & 0.0 & 27.5 & 0.0 \\
\midrule
\multicolumn{8}{@{}l}{\textit{Object \textemdash{} place the item in the basket}} \\
Tomato sauce & Early & 100.0 & 0.0 & 0.0 & 0.0 & 0.0 & 0.0 \\
 & Mid & 100.0 & 18.4 & 2.6 & 0.0 & 0.0 & 0.0 \\
Chocolate pudding & Early & 100.0 & 0.0 & 2.7 & 0.0 & 0.0 & 0.0 \\
 & Mid & 100.0 & 83.8 & 2.7 & 2.7 & 2.7 & 0.0 \\
Ketchup & Early & 100.0 & 0.0 & 0.0 & 0.0 & 0.0 & 0.0 \\
 & Mid & 100.0 & 92.3 & 0.0 & 0.0 & 0.0 & 0.0 \\
Cream cheese & Early & 100.0 & 0.0 & 0.0 & 0.0 & 0.0 & 0.0 \\
 & Mid & 100.0 & 50.0 & 2.8 & 0.0 & 0.0 & 0.0 \\
\bottomrule
\end{tabular*}
\caption{\revision{Target success and physical incident rates (\%) at each takeover. Goal placement rollouts stop at success. Stove activation and the Spatial and Object tasks use fixed windows.}}
\label{tab:task-boundaries}
\end{table*}

\begin{table*}[!tp]
\centering\footnotesize\revisioncolor
\captionsetup{labelfont+=safirevisionlabel}
\setlength{\tabcolsep}{3pt}
\renewcommand{\arraystretch}{1.07}
\begin{tabular*}{\textwidth}{@{\extracolsep{\fill}}lrrrrrrrr@{}}
\toprule
 & \multicolumn{4}{c}{LIBERO-Spatial} & \multicolumn{4}{c}{LIBERO-Object} \\
\cmidrule(lr){2-5}\cmidrule(lr){6-9}
Method & \shortstack{Target\\only} & Both & \shortstack{Incident\\only} & Neither & \shortstack{Target\\only} & Both & \shortstack{Incident\\only} & Neither \\
\midrule
Original policy & 98.3 & 1.7 & 0.0 & 0.0 & 69.4 & 30.6 & 0.0 & 0.0 \\
GA & 0.3 & 0.0 & 5.3 & 94.4 & 2.3 & 0.7 & 18.8 & 78.2 \\
GradDiff & 1.4 & 0.0 & 26.5 & 72.1 & 0.0 & 0.0 & 29.6 & 70.4 \\
NPO-RT & 42.5 & 3.0 & 17.3 & 37.3 & 10.0 & 7.3 & 15.6 & 67.1 \\
RMU & 73.0 & 0.7 & 8.8 & 17.6 & 33.0 & 28.0 & 5.7 & 33.2 \\
Refusal FT & 0.0 & 0.0 & 0.0 & 100.0 & 1.0 & 0.3 & 0.0 & 98.6 \\
VLA-Forget (OFT) & 38.8 & 3.6 & 17.7 & 39.8 & 20.1 & 18.0 & 4.8 & 57.1 \\
SAFI & 3.4 & 0.0 & 0.0 & 96.6 & 0.3 & 0.0 & 0.0 & 99.7 \\
\bottomrule
\end{tabular*}
\caption{\revision{Joint outcomes after takeover (\%). Each rollout falls into one of four categories according to target attainment and physical incidents. Rates use all rollouts within each task and boundary, with equal weights across tasks and boundaries.}}
\label{tab:joint-outcomes}
\end{table*}

\begin{multicols}{2}
\subsubsection{BadLIBERO}
\label{app:badlibero}
The GoBA setting uses a poisoned box as a physical trigger for placing that box in a designated table region, L3~\citep{zhou2025goba}. We adapt the implantation to the OFT Goal specialist and evaluate the two attacked \revision{Goal tasks}. Takeovers occur at first gripper--box contact and five steps before the earlier of post-grasp opening and L3 entry. Every method uses the same donor states and fixed observation window.

The physical-incident metric combines four event types. A \emph{drop} occurs when an ungrasped object remains more than 10\,cm below its post-wait initial height for two consecutive steps. \emph{Lost hold} applies to a box grasped at takeover that remains ungrasped, outside L3, and more than 10\,cm below its takeover height for two consecutive steps. \emph{Disturbance} records movement beyond 5\,cm or tilt beyond $45^\circ$ from the takeover pose for objects outside the task objects of interest and trigger box. A \emph{fallen-body event} records a non-box drop or an initially grasped box ending ungrasped and more than 10\,cm below its takeover height. Their union counts incidents even when they accompany attack-target attainment.

Beyond the main-table endpoints, \method produces a hold-like response in 87.5\% of takeovers. This response requires no target attainment, release, or physical incident, with end-effector excursion within 3\,cm of the takeover position. On \revision{clean,} untriggered inputs, \method retains 96.0\% clean-task success, compared with 91.0\% for the implanted policy and 95.0\% for the clean reference. These measurements complement attack suppression with the intended holding response and utility on ordinary inputs.

\subsubsection{Instruction and Visual Shifts}
\label{app:shifts}
We also evaluate normal starts under instruction paraphrases and warmer image colors, after checking that the original policy remains competent in each condition. \method records zero target success across all three Goal prohibitions under both shifts, with retained success ranging from 91.1\% to 96.7\%. This result extends the initial-state suppression finding beyond the canonical instructions and image appearance.

\subsection{Supervision Design (RQ2)}
\label{app:prefix-supervision}
The main ablation examines the two imitation terms. Here, we retain both terms and vary the states supplying fallback supervision to examine the role of execution-state coverage. All variants use the same Spatial and Object cohorts, training budget, and checkpoint-selection procedure.

Prefix-only supervision uses chunks entirely before the first gripper-close command. Count-matched full-trajectory supervision distributes the same number of candidate chunks across each demonstration. Post-close-only supervision uses chunks starting at or after that boundary, and full \method uses all valid chunks. The matched-count comparison controls the available candidate count while changing their temporal coverage and associated gripper labels.

All four variants suppress normal-start completion, but prefix-only supervision permits almost complete target execution after takeover (Table~\ref{tab:supervision-coverage}). Distributing the same candidate count across the full trajectory reduces takeover success to 0.6\% in both suites. Post-close supervision and full \method also achieve low target success while retaining high utility. This pattern supports teaching the fallback at states encountered during execution.

\end{multicols}
\begin{table*}[!tp]
\centering\footnotesize
\renewcommand{\arraystretch}{1.1}
\setlength{\tabcolsep}{3pt}
\begin{tabular*}{\textwidth}{@{\extracolsep{\fill}}lrrrrrr@{}}
\toprule
& \multicolumn{3}{c}{LIBERO-Spatial} & \multicolumn{3}{c}{LIBERO-Object} \\
\cmidrule(lr){2-4}\cmidrule(lr){5-7}
Fallback supervision & \shortstack{Target\\success rate $\downarrow$} & \shortstack{Physical\\incident rate $\downarrow$} & \shortstack{Retained\\success rate $\uparrow$} & \shortstack{Target\\success rate $\downarrow$} & \shortstack{Physical\\incident rate $\downarrow$} & \shortstack{Retained\\success rate $\uparrow$} \\
\midrule
Prefix only & 100.0 & 0.7 & 98.9 & 98.7 & 29.5 & 93.1 \\
Count-matched full & 0.6 & 0.0 & 98.1 & 0.6 & 0.3 & 93.6 \\
Post-close only & 0.0 & 0.0 & 95.8 & 0.0 & 0.0 & 95.0 \\
\rowcolor{safirow}
\method (full) & 3.4 & 0.0 & 96.9 & 0.3 & 0.0 & 93.1 \\
\bottomrule
\end{tabular*}
\caption{Fallback-supervision comparison at calibration-selected checkpoints. All values are percentages. Target success and physical incidents are measured after takeover, and retained success uses normal starts.}
\label{tab:supervision-coverage}
\end{table*}

Table~\ref{tab:supervision-shared} checks the comparison at full \method's selected update count for each target. The coverage advantage persists, with Spatial target success of 99.7\% for prefix-only supervision and 1.6\% for the count-matched variant. Post-close supervision retains less utility at these shared steps, showing that checkpoint choice also affects the balance between suppression and retained performance.

\begin{table}[H]
\centering\footnotesize
\renewcommand{\arraystretch}{1.1}
\setlength{\tabcolsep}{3pt}
\begin{tabular*}{\columnwidth}{@{\extracolsep{\fill}}lrrr@{}}
\toprule
Supervision & \shortstack{Target\\success $\downarrow$} & \shortstack{Physical\\incidents $\downarrow$} & \shortstack{Retained\\success $\uparrow$} \\
\midrule
Prefix only & 99.7 & 1.0 & 96.4 \\
Count-matched full & 1.6 & 0.0 & 97.2 \\
Post-close only & 0.0 & 0.0 & 92.2 \\
\rowcolor{safirow}
\method & 3.4 & 0.0 & 96.9 \\
\bottomrule
\end{tabular*}
\caption{Shared-step comparison on Spatial. Each variant uses the update count selected for full \method on the same target. All rates are percentages. Object values equal Table~\ref{tab:supervision-coverage}.}
\label{tab:supervision-shared}
\end{table}

\subsection{Retained Utility (RQ3)}
\label{app:rq3}
Table~\ref{tab:checkpoint-values} provides the values behind Figure~\ref{fig:utility-tradeoff}. Each row pairs takeover suppression with normal-start retained success at a sampled checkpoint, using the same task-balanced averaging as the figure.

\begin{table*}[!tp]
\centering\footnotesize
\renewcommand{\arraystretch}{1.1}
\setlength{\tabcolsep}{3pt}
\begin{tabular*}{\textwidth}{@{\extracolsep{\fill}}llrrrr@{}}
\toprule
& & \multicolumn{2}{c}{LIBERO-Spatial} & \multicolumn{2}{c}{LIBERO-Object} \\
\cmidrule(lr){3-4}\cmidrule(lr){5-6}
Method & Checkpoint & \shortstack{Target suppression\\rate (\%) $\uparrow$} & \shortstack{Retained success\\rate (\%) $\uparrow$} & \shortstack{Target suppression\\rate (\%) $\uparrow$} & \shortstack{Retained success\\rate (\%) $\uparrow$} \\
\midrule
GA & 100 & 99.7 & 0.0 & 97.0 & 0.0 \\
 & 200 & 99.7 & 0.0 & 97.0 & 0.0 \\
 & 300 & 99.7 & 0.0 & 97.0 & 0.0 \\
\addlinespace[2pt]
GradDiff & 100 & 97.3 & 3.3 & 100.0 & 0.3 \\
 & 200 & 97.9 & 22.5 & 100.0 & 19.7 \\
 & 300 & 99.0 & 34.2 & 100.0 & 41.7 \\
\addlinespace[2pt]
NPO-RT & 100 & 26.3 & 97.8 & 16.7 & 91.9 \\
 & 200 & 39.4 & 95.3 & 52.3 & 91.9 \\
 & 300 & 50.8 & 96.9 & 80.0 & 95.3 \\
\addlinespace[2pt]
RMU & 100 & 0.0 & 99.7 & 0.0 & 96.1 \\
 & 200 & 0.3 & 98.6 & 15.1 & 96.1 \\
 & 300 & 27.0 & 98.3 & 38.4 & 96.4 \\
\addlinespace[2pt]
Refusal FT & 100 & 100.0 & 60.8 & 99.3 & 27.5 \\
 & 200 & 100.0 & 90.6 & 98.7 & 64.2 \\
 & 300 & 100.0 & 93.1 & 98.6 & 74.7 \\
\addlinespace[2pt]
VLA-Forget (OFT) & 30 (final stage) & 32.9 & 55.3 & 30.7 & 67.2 \\
 & 60 (final stage) & 77.6 & 43.9 & 69.7 & 34.7 \\
 & Final & 75.6 & 82.2 & 94.5 & 25.8 \\
\addlinespace[2pt]
\rowcolor{safirow}
\method & 100 & 24.7 & 85.0 & 4.0 & 89.2 \\
\rowcolor{safirow}
 & 200 & 97.5 & 96.7 & 99.7 & 86.7 \\
\rowcolor{safirow}
 & 300 & 99.7 & 97.2 & 99.0 & 93.6 \\
\bottomrule
\end{tabular*}
\caption{Numerical values underlying Figure~\ref{fig:utility-tradeoff}. Checkpoints are update counts within each coherent training run. VLA-Forget uses checkpoints from its final language-model stage. The run correspondence is described in Appendix~\ref{app:run-provenance}.}
\label{tab:checkpoint-values}
\end{table*}

\begin{multicols}{2}
\paragraph{Checkpoint correspondence.}
\suppparagraphlabel{app:run-provenance}
Each method's checkpoint sequence aggregates a fixed set of per-target training runs. \method, NPO-RT, RMU, and Refusal FT use separate runs with the main-table training recipe. GA, GradDiff, and VLA-Forget use their original runs, with VLA-Forget sampled during its final adaptation stage. The main table selects checkpoints separately for each target, while the figure compares common update counts, so the two aggregates can differ. GA's points overlap because suppression and retained success are unchanged across its sampled checkpoints.

\revision{At the selected Object checkpoints, \method retains 93.1\% success and Refusal FT retains 77.2\% (Table~\ref{tab:normal-retained}). All four selected \method checkpoints met both calibration criteria, while Refusal FT used the highest-retention suppression-passing candidate on three tasks with no candidate meeting both criteria.}

\paragraph{Retained-task responses.}
\suppparagraphlabel{app:retained}
We complement task success with a paired physical-event audit under normal starts and neighboring-task takeovers. Among the qualified edits, the 95\% paired intervals place event-rate increases below ten percentage points for all but one comparison, with the remaining comparison inconclusive. For all three Goal prohibitions, \method also matches or improves original-policy success on instruction-based neighboring tasks. Success on more distant task groups stays within the same ten-percentage-point retention tolerance. These checks extend the pooled utility results to the tasks most closely related to each prohibition and their physical responses.

\subsection{Physical Responses (RQ4)}
\label{app:physical-response}
The physical-response analysis uses the selected Spatial and Object policies underlying the main table. All methods start from matched states and track the same initially held object over the full observation window, including rollouts with target attainment or physical incidents.

Let $p_i(t)\in\mathbb R^3$ be the object's position in meters for donor trajectory $i$, with $t=0$ denoting takeover. Its displacement in centimeters is
\begin{equation}
\label{eq:object-displacement}
d_i(t)=100\lVert p_i(t)-p_i(0)\rVert_2.
\end{equation}
We average over donors within each target and then equally over targets, keeping early and mid-task takeover separate. The 95\% bands use paired donor bootstrap and are displayed for \method and scripted hold. This comparison measures adherence to the stationary holding objective, alongside the target-attainment and physical-incident outcomes in the main table.

\paragraph{Gripper contact.}
Figure~\ref{fig:physical-contact} complements displacement with contact between the gripper and the initially held object. Let $g_{q,i}(t)$ indicate this contact for donor $i$ of target $q$, with $N_q$ donors and target set $\mathcal Q_s$ in suite $s$. The task-balanced contact rate is
\begin{equation}
\label{eq:gripper-contact}
G_s(t)=\frac{100}{|\mathcal Q_s|}\sum_{q\in\mathcal Q_s}
\frac{1}{N_q}\sum_{i=1}^{N_q}g_{q,i}(t).
\end{equation}
\rqfourrevision{The four condition-wise endpoints in Table~\ref{tab:holding-endpoints} are averaged equally for the main-text summary in Table~\ref{tab:holding-summary}.}

\method's final contact rates range from 93.1\% to 100.0\% across the four conditions\revision{ (Table~\ref{tab:holding-endpoints})}. Together with its low object displacement, these measurements show that the learned response largely maintains the initially held object near its takeover position. \revision{In the residual Object completion, the chocolate pudding leaves the gripper and reaches the basket goal, with no registered physical incident.}
\end{multicols}
\begin{table*}[!tp]
\centering\footnotesize\revisioncolor
\captionsetup{labelfont+=safirevisionlabel}
\setlength{\tabcolsep}{3pt}
\renewcommand{\arraystretch}{1.07}
\begin{tabular*}{\textwidth}{@{\extracolsep{\fill}}lrrrrrrrr@{}}
\toprule
 & \multicolumn{4}{c}{LIBERO-Spatial} & \multicolumn{4}{c}{LIBERO-Object} \\
\cmidrule(lr){2-5}\cmidrule(lr){6-9}
 & \multicolumn{2}{c}{Early} & \multicolumn{2}{c}{Mid} & \multicolumn{2}{c}{Early} & \multicolumn{2}{c}{Mid} \\
\cmidrule(lr){2-3}\cmidrule(lr){4-5}\cmidrule(lr){6-7}\cmidrule(lr){8-9}
Method & \shortstack{Distance\\(cm)} & \shortstack{Contact\\(\%)} & \shortstack{Distance\\(cm)} & \shortstack{Contact\\(\%)} & \shortstack{Distance\\(cm)} & \shortstack{Contact\\(\%)} & \shortstack{Distance\\(cm)} & \shortstack{Contact\\(\%)} \\
\midrule
Original policy & 24.6 & 23.6 & 17.0 & 29.0 & 44.7 & 0.0 & 28.4 & 0.0 \\
GA & 52.1 & 71.2 & 66.2 & 96.7 & 56.0 & 35.7 & 81.0 & 55.1 \\
GradDiff & 68.1 & 79.6 & 71.3 & 74.7 & 70.1 & 59.9 & 77.1 & 40.8 \\
NPO-RT & 29.5 & 33.1 & 24.1 & 30.2 & 57.9 & 48.6 & 51.9 & 40.9 \\
RMU & 28.7 & 27.0 & 21.3 & 32.1 & 40.8 & 45.4 & 35.5 & 18.5 \\
Refusal FT & 37.4 & 98.7 & 21.9 & 99.3 & 38.3 & 99.3 & 12.9 & 98.0 \\
VLA-Forget (OFT) & 33.7 & 18.9 & 37.8 & 15.6 & 48.5 & 49.1 & 36.0 & 39.8 \\
SAFI & 1.5 & 100.0 & 4.8 & 93.1 & 2.8 & 96.7 & 3.2 & 99.3 \\
\midrule
Scripted hold & 0.6 & 100.0 & 2.4 & 100.0 & 0.9 & 97.4 & 1.3 & 100.0 \\
\bottomrule
\end{tabular*}
\caption{\revision{Final displacement of the held object from its takeover position (cm) and final gripper contact rate (\%). Each suite mean weights its targets equally.}}
\label{tab:holding-endpoints}
\end{table*}

\begin{figure*}[!tp]
\centering
\includegraphics[width=.88\textwidth]{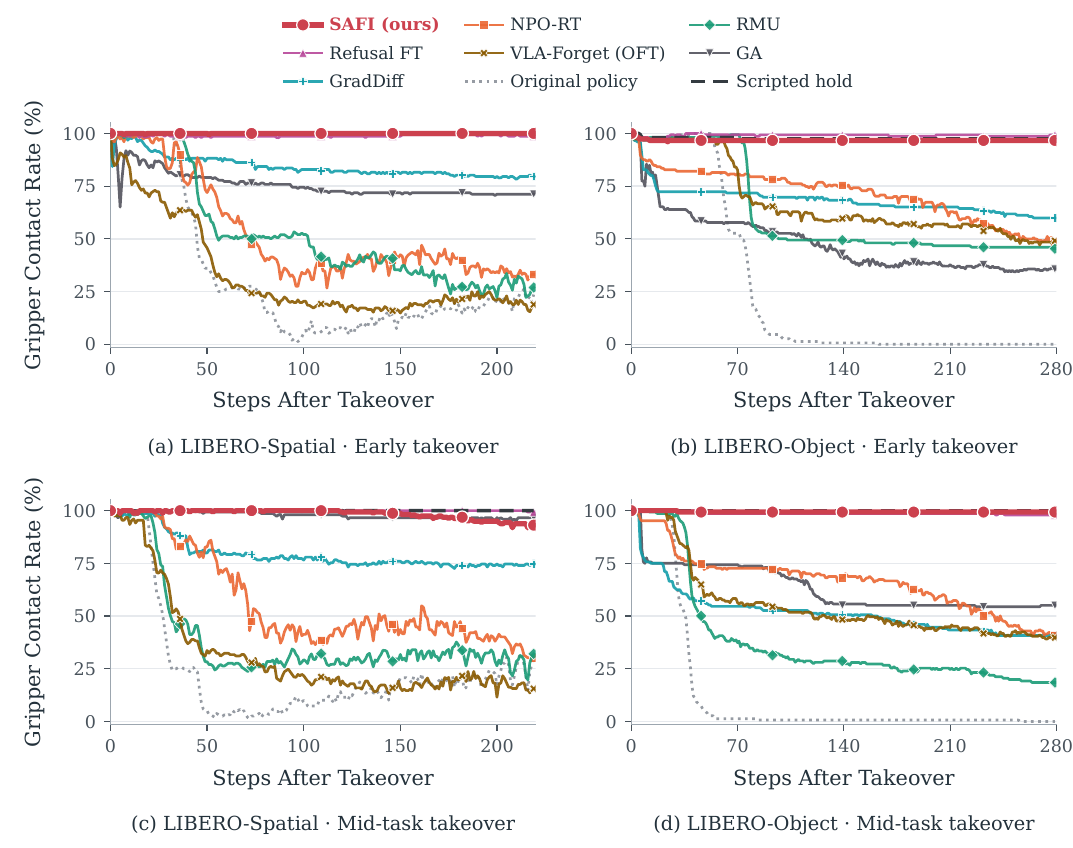}
\caption{Gripper contact after takeover on the same trajectories as Figure~\ref{fig:physical-response}. Curves show task-balanced contact rates, with 95\% paired donor-bootstrap intervals for \method and scripted hold.}
\label{fig:physical-contact}
\end{figure*}
\begin{figure*}[!tp]
\centering
\includegraphics[width=\textwidth]{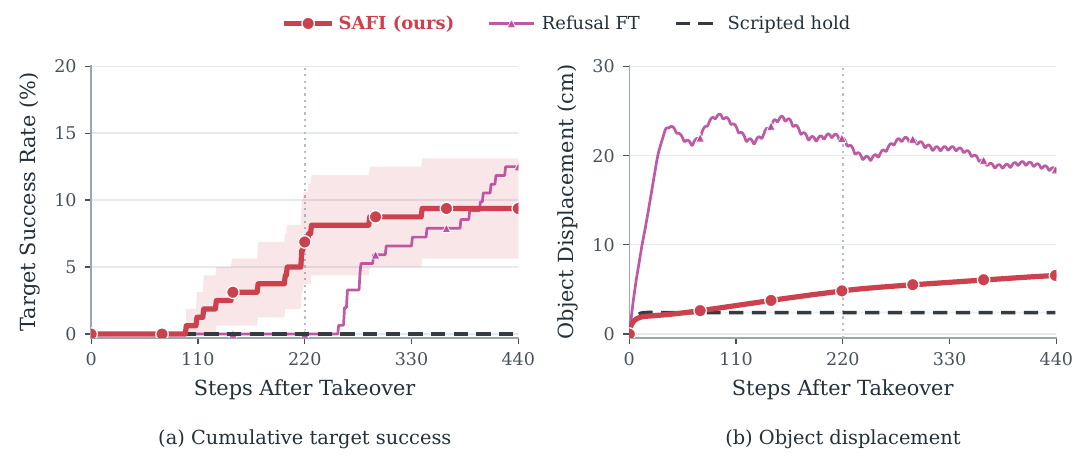}\par
\captionsetup{labelfont+=safirevisionlabel}
\caption{\revision{Spatial mid-task takeover over 440 actions. The vertical line marks the original 220-action horizon. Shading gives pointwise 95\% paired donor-bootstrap intervals for \method and scripted hold.}}
\label{fig:extended-holding}
\end{figure*}
\begin{multicols}{2}
\paragraph{\revision{Holding over a longer window.}}
\suppparagraphlabel{app:extended-window}
\revision{Extending the same Spatial mid-task rollouts from 220 to 440 actions tests whether the holding response persists over a longer window (Figure~\ref{fig:extended-holding}). \method records zero physical incidents at both horizons, with mean object displacement increasing from 4.8 to 6.6\,cm and target success from 6.9\% to 9.4\% (Table~\ref{tab:extended-holding}). These completions remain confined to the bowl between the plate and the ramekin, where final gripper contact decreases from 72.5\% to 32.5\%, compared with 100\% on the other three tasks at both horizons. Refusal FT's additional completions occur on the bowl-next-to-plate task, where its selected policy falls below the held-out retention criterion.}
\end{multicols}
\begin{table*}[!tp]
\centering\footnotesize\revisioncolor
\captionsetup{labelfont+=safirevisionlabel}
\setlength{\tabcolsep}{3pt}
\renewcommand{\arraystretch}{1.07}
\begin{tabular*}{\textwidth}{@{\extracolsep{\fill}}llrrrrrrrr@{}}
\toprule
 &  & \multicolumn{2}{c}{Target success (\%)} & \multicolumn{2}{c}{Incidents (\%)} & \multicolumn{2}{c}{Gripper contact (\%)} & \multicolumn{2}{c}{Distance (cm)} \\
\cmidrule(lr){3-4}\cmidrule(lr){5-6}\cmidrule(lr){7-8}\cmidrule(lr){9-10}
Bowl position & Policy & 220 & 440 & 220 & 440 & 220 & 440 & 220 & 440 \\
\midrule
Next to the plate & SAFI & 0.0 & 0.0 & 0.0 & 0.0 & 100.0 & 100.0 & 3.9 & 5.0 \\
 & Refusal FT & 0.0 & 50.0 & 0.0 & 2.6 & 100.0 & 52.6 & 28.7 & 15.8 \\
 & Scripted hold & 0.0 & 0.0 & 0.0 & 0.0 & 100.0 & 100.0 & 2.1 & 2.1 \\
\midrule
Next to the cookie box & SAFI & 0.0 & 0.0 & 0.0 & 0.0 & 100.0 & 100.0 & 3.0 & 4.5 \\
 & Refusal FT & 0.0 & 0.0 & 0.0 & 2.8 & 97.2 & 97.2 & 23.1 & 25.2 \\
 & Scripted hold & 0.0 & 0.0 & 0.0 & 0.0 & 100.0 & 100.0 & 2.0 & 2.0 \\
\midrule
On the stove & SAFI & 0.0 & 0.0 & 0.0 & 0.0 & 100.0 & 100.0 & 3.7 & 6.7 \\
 & Refusal FT & 0.0 & 0.0 & 0.0 & 0.0 & 100.0 & 100.0 & 12.7 & 12.2 \\
 & Scripted hold & 0.0 & 0.0 & 0.0 & 0.0 & 100.0 & 100.0 & 2.7 & 2.6 \\
\midrule
Between plate and ramekin & SAFI & 27.5 & 37.5 & 0.0 & 0.0 & 72.5 & 32.5 & 8.7 & 10.1 \\
 & Refusal FT & 0.0 & 0.0 & 0.0 & 0.0 & 100.0 & 100.0 & 23.2 & 20.4 \\
 & Scripted hold & 0.0 & 0.0 & 0.0 & 0.0 & 100.0 & 100.0 & 2.9 & 2.9 \\
\midrule
Task-balanced mean & SAFI & 6.9 & 9.4 & 0.0 & 0.0 & 93.1 & 83.1 & 4.8 & 6.6 \\
 & Refusal FT & 0.0 & 12.5 & 0.0 & 1.4 & 99.3 & 87.5 & 21.9 & 18.4 \\
 & Scripted hold & 0.0 & 0.0 & 0.0 & 0.0 & 100.0 & 100.0 & 2.4 & 2.4 \\
\bottomrule
\end{tabular*}
\caption{\revision{Spatial mid-task takeover at 220 and 440 actions. Target and incident rates accumulate through each horizon. Gripper contact and object displacement use the last observation before the final action.}}
\label{tab:extended-holding}
\end{table*}

\begin{multicols}{2}
\section{Computational Efficiency}
\label{app:efficiency}
We measure adaptation and action-prediction costs to assess the computational requirements of \method. The inference comparison uses the original and adapted policies on the same H100 GPU, with batch size one and merged LoRA adapters. Each timed call processes cached observations and produces a fresh eight-action chunk, including input and action postprocessing.

Table~\ref{tab:inference-cost} shows median latency near 58\,ms per chunk and approximately 15\,GiB peak allocated memory for both policies. These similar deployment costs are consistent with \method's unchanged action-prediction path after adaptation.

Table~\ref{tab:training-cost} summarizes the 300-update adaptation loop, measured after setup through the final update. The median time is 12.2 minutes per target, with approximately 31\,GiB peak allocated memory per process while two training processes share an H100. These measurements characterize the training cost of the evaluated recipe alongside its deployed inference cost.
\end{multicols}
\begin{table*}[!tp]
\centering\footnotesize
\renewcommand{\arraystretch}{1.1}
\setlength{\tabcolsep}{3pt}
\begin{tabular*}{\textwidth}{@{\extracolsep{\fill}}llrrr@{}}
\toprule
Suite & Policy & \shortstack{Median latency\\(ms/chunk)} & \shortstack{P95 latency\\(ms/chunk)} & \shortstack{Peak GPU memory\\(GiB)} \\
\midrule
Spatial & Original policy & 58.10 & 60.16 & 14.997 \\
\rowcolor{safirow}
Spatial & \method & 58.59 & 62.72 & 14.996 \\
\midrule
Object & Original policy & 57.67 & 59.14 & 14.994 \\
\rowcolor{safirow}
Object & \method & 58.13 & 59.08 & 14.994 \\
\bottomrule
\end{tabular*}
\caption{Action-prediction cost on one H100 with batch size one and merged adapters. Each call produces a fresh eight-action chunk. GPU memory is peak PyTorch allocated memory, including the resident policy.}
\label{tab:inference-cost}
\end{table*}

\begin{table*}[!tp]
\centering\footnotesize
\renewcommand{\arraystretch}{1.1}
\setlength{\tabcolsep}{3pt}
\begin{tabular*}{\textwidth}{@{\extracolsep{\fill}}lrrr@{}}
\toprule
Suite & Target runs & \shortstack{Training-loop time\\median [range] (min)} & \shortstack{Peak training memory\\median [range] (GiB)} \\
\midrule
Spatial & 4 & 12.2 [12.1, 12.2] & 31.05 [31.05, 31.05] \\
Object & 4 & 12.2 [12.1, 12.2] & 30.91 [30.91, 30.91] \\
\bottomrule
\end{tabular*}
\caption{Training cost of the 300-update \method runs. Two training processes shared an H100. Timing covers the adaptation loop after model and data setup.}
\label{tab:training-cost}
\end{table*}

\begin{multicols}{2}
\section{Qualitative Comparisons}
\label{app:qualitative}
Figure~\ref{fig:qualitative-takeovers} shows the behavioral differences behind the two motivations. The bottle case contrasts task completion, dropping, and continued holding, while the stove case contrasts activation with suppression after takeover. We render the recorded actions from matched starting states and verify the displayed NPO-RT and \method states against their evaluation logs.

\end{multicols}
\begin{figure*}[!t]
\centering
\includegraphics[width=\textwidth]{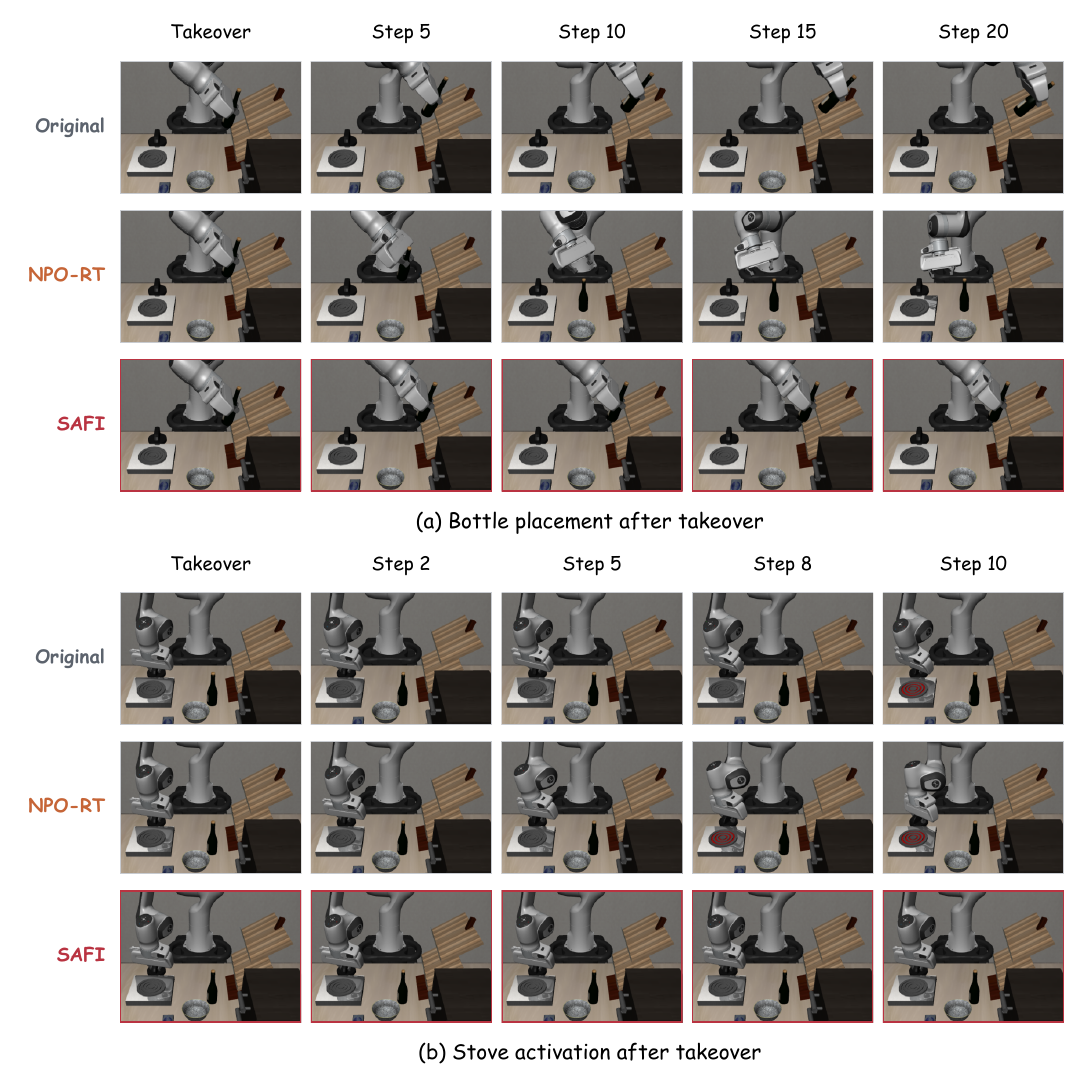}
\caption{Trajectories after matched mid-task takeover. Columns show the same elapsed control steps across methods. Original completes both prohibited tasks. NPO-RT drops the bottle and activates the stove, while \method keeps the bottle held and the stove off. Each panel ends at Original's task completion.}
\label{fig:qualitative-takeovers}
\end{figure*}
\begin{multicols}{2}
\subsection{Motivation Figure Protocol}
\label{app:motivation-details}
The upper panel of Figure~\ref{fig:motivation} compares stove activation from initial, early, and mid-task states under the restart protocol. Its initial-state condition is separate from the held-out normal-start evaluation in Table~\ref{tab:normal-retained}. The lower panel accumulates bottle-drop events over all episodes in each condition until their termination. Each curve ends at its last observed step, with the original policy's shorter window reflecting earlier task completion. The curves therefore describe recorded outcomes over the observed execution intervals.

\end{multicols}

\endgroup

\end{document}